\documentclass[11pt]{article}

\usepackage[utf8]{inputenc}
\usepackage[T1]{fontenc}
\usepackage{csquotes}
\usepackage[style=authoryear,giveninits=true,url=false,isbn=false,doi=false,related=false, natbib,maxbibnames=9,uniquename=false]{biblatex}
\usepackage[french, main=english]{babel}
\usepackage{geometry}
\usepackage{booktabs}
\usepackage{xcolor}
\usepackage{amsthm}
\usepackage{amsmath}
\usepackage{amssymb}
\usepackage{extarrows}
\usepackage{mathrsfs}
\usepackage{mathtools}
\usepackage{stmaryrd}
\usepackage{breakcites}
\usepackage{stackrel}
\usepackage[hang]{footmisc}
\usepackage{setspace}
\usepackage[nottoc, notlof, notlot]{tocbibind}
\usepackage{import}
\usepackage[explicit]{titlesec}

\usepackage{style}

\renewcommand{\cite}{\citet}

\usepackage[breaklinks]{hyperref}
\hypersetup{linkbordercolor = {pretty_red}}

\DeclareMathOperator*{\argmin}{\mathrm{argmin}}

\newcommand{\myfont}[1]{{\usefont{T1}{ubuntu-bold}{m}{n} #1}}
\titleformat{\section}[block]{\sffamily\centering\textbf{\thesection. \, \MakeUppercase{#1}}}{}{1em}{}
\titleformat{\subsection}[block]{\sffamily\textbf{\thesubsection. \, #1}}{}{1em}{}

\newcommand{\footnoteremember}[2]{
    \footnote{#2}
    \newcounter{#1}
    \setcounter{#1}{\value{footnote}}
}
\newcommand{\footnoterecall}[1]{
    \footnotemark[\value{#1}]
}

\date{}

\begin{document}

\vspace{-2cm}
\begin{flushleft}
    \Huge\myfont{Sampling From the Wasserstein Barycenter}
\end{flushleft}

\vspace{0.25cm}

\noindent
\textbf{Chiheb Daaloul}\footnoteremember{ecm}{Aix-Marseille Univ., CNRS, I2M, UMR7373, Centrale Marseille, 13451 Marseille, France}
\textbf{Thibaut Le Gouic}\footnoteremember{mit}{
    Massachusetts Institute of Technology, 
    Department of Mathematics, 
    77 Massachusetts Avenue, Cambridge, MA 02139-4307, USA
}
\textbf{Jacques Liandrat}\footnoterecall{ecm}
\textbf{Magali Tournus}\footnoterecall{ecm}
\vspace{1mm}

\vspace{-5mm}
\renewcommand{\abstractname}{}
\begin{abstract}
    \noindent
    \textit{Abstract. }
    This work presents an algorithm to sample from the Wasserstein barycenter of absolutely continuous measures. 
    Our method is based on the gradient flow of the multimarginal formulation of the Wasserstein barycenter, with 
    an additive penalization to account for the marginal constraints.
    We prove that the minimum of this penalized multimarginal formulation is achieved for a coupling that is 
    close to the Wasserstein barycenter.
    The performances of the algorithm are showcased in several settings.
\end{abstract}

\section{Introduction}

The barycenter in the space of probability measures equipped with the Wasserstein distance, first 
introduced by \cite{agueh_carlier_wbary} gained a lot of popularity in recent years.
The most striking property of the Wasserstein space is probably how its geodesics can be interpreted as a displacement of 
particles in $\RR^d$, allowing the Wasserstein barycenter of measures to take into account the geometry of the 
underlying space $\RR^d$.
Thanks to its geometric interpretation, it proved useful in a variety of applications.
For instance, it has been applied in image processing:
\cite{julien_texture_mixing} developed an algorithm for
texture synthesis built around this measure, \cite{barre_atm_wbary} used the
Wasserstein barycenter of images to improve the precision of gas emission sourcing,
and it played a central role in \cite{gramfort_foptand} to compare brain image data. 
In the field of fairness, \cite{barrio_fairness} and \cite{legouic_fairness} leveraged
the distribution to tackle the problem of fairness in regression and classification problems.
Barycenters also found applications in Bayesian inference; \cite{srivastava2015wasp,bayes_inference_example}
proposed a method to accelerate the computation of posterior distributions for Bayesian inference 
using the Wasserstein barycenter.

More formally, let us denote by 
$(P_2(\RR^d),W_2)$ the set of all measures defined on $\RR^d$ with finite second order moment, endowed with the Wasserstein distance
\begin{equation*}
    W_2 : (\mu, \nu) \longmapsto 
        \sqrt{\inf_{\gamma} \int_{\RR^d \times \RR^d} \nabs{x-y}^2 \Di\gamma(x,y)} \,,
\end{equation*}
where the infimum is taken over the set of couplings of $\mu, \nu$.
A Wasserstein barycenter $b$ of $\mu_1, \dots, \mu_n$ with weights $\lambda_1, \dots, \lambda_n$
is a minimizer of the map $\nu \mapsto \sum \lambda_i W_2^2(\mu_i, \nu)$,
\ie the measure $b$ corresponds to a Fréchet mean of $\mu_1, \dots, \mu_n$ in the Wasserstein
space $(P_2(\RR^d), W_2)$.
Equivalently, the multimarginal formulation of the barycenter
problem asserts that $b$ is obtained by pushing forward the minimizer $\gamma^\star$ of 
a functional $G$ (see equation \eqref{eqn:G_functional}) defined over the set of couplings of the 
$\mu_i$'s. Under mild conditions on the $\mu_i$'s,
both problems admit unique solutions, and yield the same measure.
Section \ref{sect:notions} provides more details on the Wasserstein barycenter.

Most methods in the literature propose to estimate the Wasserstein barycenter with a discrete measure \eg
\cite{cuturi2014fast,solomon2015convolutional,benamou2015iterative}.
However, this approach does not scale well with the dimension as noted by \cite{altschuler2021wasserstein}.
Indeed, they showed that computing the infimum of $\nu \mapsto \sum \lambda_i W_2^2(\mu_i, \nu)$ when the
$\mu_i$ are discrete measures is a NP-hard problem in the dimension, even
when only approximate solutions are acceptable.
We can, however, avoid estimating the density altogether
and focus on generating samples distributed according to the barycenter of known measures.
Given the broad applicability of the Wasserstein barycenter and of sampling techniques in general,
we believe that such sampling procedures should be part of the statistician's toolbox.

We are motivated by the success of sampling methods in high dimensional settings, which is
due to the possibility of integrating functions against the target measure without resorting to discretization on a large grid.
Markov chain methods have become popular tools to this end. 
The most well known among them are probably Hamiltonian Monte Carlo methods, variants of the 
Metropolis-Hastings algorithm relying on simulations to generate diverse samples from a distribution
(\cite[Chapter 11]{bishop_prml}), and Langevin diffusion
(\cite[Chapter 4]{pavliotis_stochastic_processes_and_apps_2015}), which transports
points along random trajectories to redistribute them according to the target measure. Such transportation
methods are computationally attractive since one only needs to store minimal information about how to
move the particles at each iteration. 

In their celebrated work, \cite{jko} showed that the marginals of Langevin
diffusion are distributed according to gradient flows in the Wasserstein space of the Kullback-Leibler divergence with
respect to the stationary measure. This insight, studied rigorously in \cite{ags}, brought to light a
connection between sampling and optimization which added a new perspective to the study of Monte Carlo
algorithms (\eg \cite{vempala_rates_ula} analyze the Unadjusted Langevin Algorithm and 
\cite{chewi2020exponential} analyze the Mirror Langevin diffusion in this framework). 
Inspired by this insight, we aim to minimize the multimarginal
formulation $G$ with gradient descent in the Wasserstein space. This procedure, like
Langevin diffusion, iteratively redistributes randomly initialized points to produce a sample from
the barycenter. 
We therefore recover an approximation of the optimal 
coupling through the samples, which can be desirable in applications. 

However since the barycenter is defined by a constrained optimization problem,
one cannot expect that the constraints will remain satisfied along the gradient flow.
This issue of constraints has been tackled in a variety of ways in the literature. A natural 
solution consists in restricting the domain of possible directions at each iteration,
which hints to the Frank-Wolfe algorithm, \ie to choose the steepest descent direction
in the admissible set.
This approach was studied in \cite{luise2019sinkhorn} within the
context of regularized optimal transport, where the regularized Wasserstein barycenter (also known
as the Sinkhorn barycenter) minimizes a sum of Sinkhorn divergences. 
When the admissible set of directions forms a Reproducing Kernel Hilbert Space (RKHS) with suitable kernel, 
\cite{sinkhorn_bary} propose to generate samples distributed 
according to the Sinkhorn barycenter by iterative pushforward of an initial measure with the map 
$\id_{\RR^d} - h \cdot d$, where $h$ is a step size and $d$ is the direction of steepest descent in the RKHS.
Both methods operate on discrete measures and the authors prove that the
continuous measure is recovered as a (weak) limit when the number of samples increases to ensure consistency.
We refer to \cite{peyre_cot} for details about regularized optimal transport.
We note that in some settings the issue of constraints can be addressed more easily, for example in the Bures-Wasserstein 
manifold (\ie the subspace of Gaussian measures in the Wasserstein space)
where the barycenter problem reduces to a finite dimensional optimization problem.
\cite{chewi_bures} propose a 
gradient descent algorithm for the original formulation of the barycenter problem.
However, we obviously cannot hope that similar properties hold for arbitrary continuous measures.

In this paper, we perform gradient descent on a penalized functional
$F^\al : P_2((\RR^d)^n) \to \RR^+$ obtained from $G$ by adding a penalization term to control the distance 
between the coupling marginals and the $\mu_i$'s.
We weigh the penalization with a coefficient $\al$ and control the induced error with $\al$. Taking advantage of
the differential calculus on the Wasserstein space, we can define a gradient flow for $F^\al$. 
To implement this procedure, we focus on the popular \ssf{SVGD} algorithm introduced in \cite{liu_wang_svgd}
to approximate the gradient of the penalization; technical details are given in Section \ref{sect:simulations}.

\begin{figure}[ht]
    \centering
    \vspace*{-1em}
    \includegraphics[ width=\textwidth]{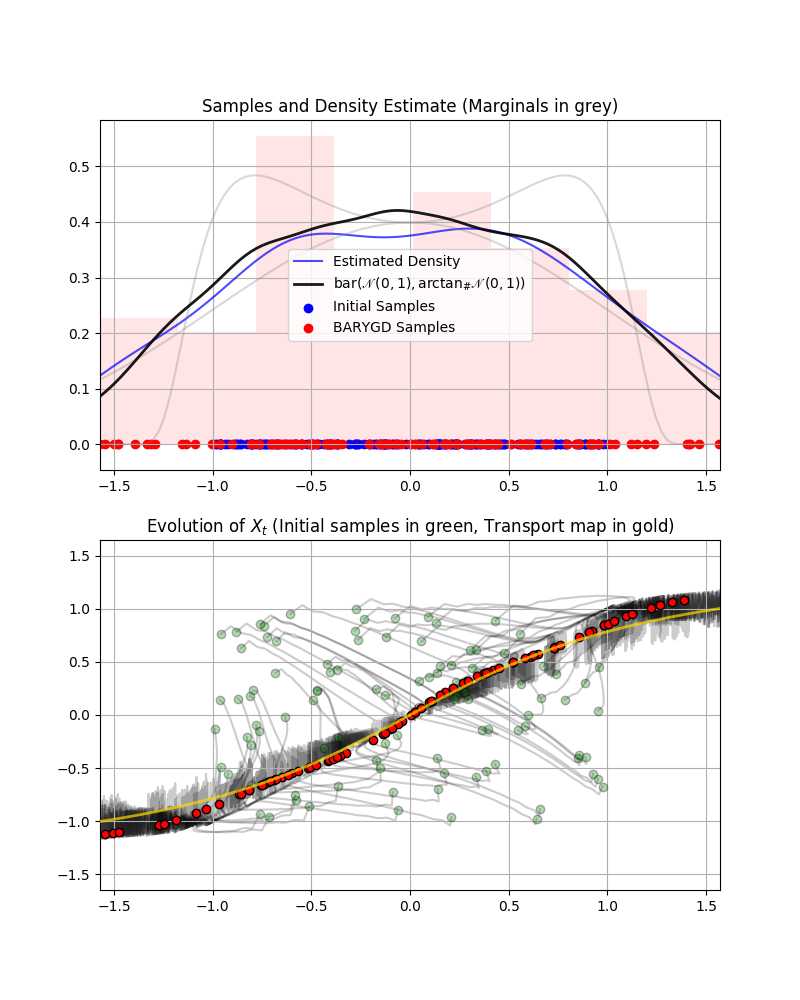}
    \vspace{-2em}
    \caption{\small Top plot: 200 samples drawn from $\bary{\prN(0,1), \arctan \pf \prN(0,1)}$ using
    \ssf{BARYGD} with a Gaussian kernel and initial step size $h = 10^{-3}$, and the resulting kernel 
    density estimate. Bottom plot: Trajectories of the samples evolving over 220 iterations, from green to
    red samples. We can see the $\arctan$ transport map is approximately reconstructed.}
    \label{fig:norm_arctan}
\end{figure}

\paragraph{Contributions.}

Inspired by the work of Jordan, Kinderlehrer and Otto cited above, we introduce a new sampling 
algorithm, called \ssf{BARYGD}, 
that performs kernelized gradient descent on a well chosen functional $F^\al$ built as a penalized version of 
the now classical multimarginal formulation of the barycenter problem introduced by \cite{agueh_carlier_wbary}.
We show that our method is consistent: with fixed penalization strength $\al$, if gradient descent yields 
a coupling $\geal$ such that $F^\al(\geal) - \inf F^\al \leq \epsilon$, then we get a quantitative bound on 
the Wasserstein distance between the approximate barycenter obtained from $\geal$ and the true barycenter 
obtained from the optimal coupling $\gamma^\star$.
This bound vanishes when $\epsilon \to 0$ and $\al \to \infty$.
As a consequence, we show that the minimizers of $F^\al$ converge to
$\gamma^\star$ when $\al \to \infty$ and quantify the rate of convergence. Furthermore, we 
perform numerical experiments with the algorithm in several settings (see Figures \ref{fig:norm_arctan} and \ref{fig:norm}).

\paragraph{Organization.}

In Section \ref{sect:notions}, we describe the Wasserstein barycenter in more details.
We present and analyze our method in Section \ref{sect:algorithm}.
Section \ref{sect:simulations} is devoted to implementation details. In Section
\ref{sect:discussion}, we discuss some open questions. 

\paragraph{Notation.}

\emph{All probability measures are assumed to be absolutely
continuous with respect to the Lebesgue measure and, with a slight abuse of notation, we 
will identify measures with their densities.}
We write $X\sim \nu$ when random variable $X$ has distribution $\nu$.
For $\mu, \nu \in P_2(\RR^d)$ and a measurable map $f : \RR^d \to \RR^d$,
we write $\nu = f \pf \mu$ if $f(X) \sim \nu$ when $X \sim \mu$. 
The set of couplings of $\mu$ and $\nu$, \ie probability measures on $\RR^d \times \RR^d$ with marginals
$\mu$ and $\nu$, is denoted by $\Pi(\mu, \nu)$.
We denote marginals by indices so $\gamma_1 = \mu$ and $\gamma_2 = \nu$ for $\gamma \in \Pi(\mu, \nu)$;
the notation extends naturally to couplings of $n$ measures.
We denote Euclidean gradients by $\nabla$
and write $\nabla_W$ for gradients in the Wasserstein space. Finally, we write indifferently $c_t = c(t)$ 
and $c' = \dot{c} = \partial_t c$ when $c : [a,b] \to X$ is a curve in some space $X$.
\section{Wasserstein barycenters}
\label{sect:notions}

This section briefly presents background notions on the Wasserstein space and its barycenters that were first 
introduced in the seminal paper of \cite{agueh_carlier_wbary}.
We refer to \cite{legouic_existence} for questions of existence and stability of the barycenter.

The Wasserstein space over $\RR^d$ is defined as the set $P_2(\RR^d)$ of probability measures over $\RR^d$ 
with finite second order moment, endowed with the distance $W_2$ defined by
\[
    W_2^2(\mu_0,\mu_1) 
        = \inf_{\gamma \in \Pi(\mu_0, \mu_1)} \int \nabs{x-y}^2\Di\gamma(x,y), \quad \forall \mu_0, \mu_1 \in P_2(\RR^d)
\]
where the infimum is taken over the set $\Pi(\mu_0, \mu_1)$ of all couplings $\gamma$ of $\mu_0$ and $\mu_1$.
The Wasserstein space is a geodesic space: for every $\mu_0$ and $\mu_1$ in $P_2(\RR^d)$ there exists 
a path $(\mu_t)_{t\in [0,1]}$ such that 
\[
    W_2(\mu_s,\mu_t) = |s-t| \, W_2(\mu_0,\mu_1), \quad \forall s,t \in [0,1].
\]
Such a path $(\mu_t)_{t\in[0,1]}$ is called a constant-speed \emph{geodesic} between its end points $\mu_0$ and $\mu_1$ 
--- we will often shorten it to \emph{geodesic}.
A functional $G$ defined over $P_2(\RR^d)$ is said to be \emph{geodesically convex} if it is convex along every geodesic.

Let $\mu_1,\dots,\mu_n\in P_2(\RR^d)$ and let $\lambda_1,\dots,\lambda_n$ be strictly positive weights 
such that $\sum \lambda_i = 1$. 
A Wasserstein barycenter of $(\mu_i)$ with weights $(\lambda_i)$ is defined as any measure
\begin{equation}
    \label{eqn:bary_def}
    b \in \argmin_{\nu \in P_2(\RR^d)} \sum \lambda_i W_2^2(\mu_i, \nu),
\end{equation}
\ie a barycenter corresponds to a Fréchet mean of the $\mu_i$ in the metric space $(P_2(\RR^d), W_2)$.
Whenever one of $\mu_1, \dots, \mu_n$ has density with respect to the Lebesgue measure, the barycenter is
unique and always defined. We assume throughout that the $\mu_i$ are absolutely continuous with
respect to the Lebesgue measure on $\RR^d$.

The minimization problem \eqref{eqn:bary_def} is equivalent to the following multimarginal problem.
Let 
$T(x) = \sum \lambda_i x_i$ for any $x \in (\RR^d)^n$ and recall that $\Pi(\mu_1, \dots, \mu_n)$ denotes the set of all couplings of 
the $\mu_i$.
The infimum of the functional
\begin{equation}
    \label{eqn:G_functional}
    G : \gamma \longmapsto \int \sum \lambda_i \nabs{x_i - T(x)}^2 \Di\gamma(x)
\end{equation}
over $\Pi(\mu_1, \dots, \mu_n)$ is achieved for a coupling $\gamma^\star$ such that
\begin{subequations}
    \begin{align}
        & b = \bary{\gamma^\star} \label{eqn:bary_pf}, \\
        & G(\gamma^\star) = \sum \lambda_i W_2^2(\mu_i, \bary{\gamma^\star}) \label{eqn:multmargform}.
    \end{align}
\end{subequations}
Moreover, if one of the $\mu_i$'s is absolutely continuous with respect to the Lebesgue measure, then $\gamma^\star$ is unique.
Note that by introducing the probability measure $P = \sum \lambda_i \delta_{\mu_i}$ on $P_2(\RR^d)$, the right hand side 
of \eqref{eqn:multmargform} corresponds to the variance of $P$.
With this notations, $b$ minimizes
the variance functional $\nu \mapsto \sum \lambda_i W_2^2(\mu_i, \nu)$ of $P$ and
$b$ corresponds to the barycenter of $P$.

In the next section, we leverage the smooth structure of the Wasserstein space in order to develop an optimization scheme 
for $G$ that will be the base of our sampling algorithm.

\section{Sampling with gradient flows}
\label{sect:algorithm}
This section defines the flow gradient of a functional on the Wasserstein space.
Some useful details of this technique introduced in \cite{jko} are gathered
in Appendix~\ref{sect:sosws}.

\subsection{Definition of the problem}

We aim to optimize the functional $G$ defined in Section \ref{sect:notions} 
with a gradient flow on the Wasserstein space. For simplicity, we first define the cost function $c$ by
\begin{equation*}
    c : x \mapsto \sum \lambda_i \nabs{x_i - T(x)}^2
\end{equation*}
then, for any $\gamma \in P_2((\RR^d)^n)$, we have
\begin{equation*}
    G(\gamma) = \int c \Di\gamma.
\end{equation*}

The set of marginal constraints $\Pi(\mu_1, \dots, \mu_n)$ is convex
in $L^2((\RR^d)^n)$ but is not geodesically convex in the Wasserstein space, which makes the multimarginal 
problem a difficult non-convex optimization problem on $P_2(\RR^d)$.
Since a gradient descent scheme on $G$ will inevitably leave the constraint set $\Pi(\mu_1, \dots, \mu_n)$, we modify the problem by enforcing 
the constraints with a penalization term that ensures that the marginals are close to the $\mu_i$'s in $\Penalization$ divergence.
We recall that the $\Penalization$ divergence between two measures $\mu, \nu$ is given by
\[
    \Penalization_\nu(\mu) = \int \bp{\frac{\mu}{\nu} - 1}^2 \Di\nu
        \textrm{  if } \mu \ll \nu \quad\textrm{and}\quad
    \Penalization_\nu(\mu) = +\infty \textrm{  otherwise}.
\]
The problem thus becomes to minimize the functional $F^\al$ defined by
\begin{equation*}
    \gamma \mapsto F^\al(\gamma) := \int c \Di\gamma + \al \sum \lambda_i \Penalization_{\mu_i}(\gamma_i),
\end{equation*}
for $\al\in[0,\infty]$ (we use the convention $0\times\infty=0$).

For $\al$ large enough, the minimizer of $F^\al$ should be close to the minimizer of $G$ with the marginal 
constraints --- which corresponds also to the minimizer of $F^\infty$.
The following proposition ensures that the minimizer exists.
\begin{proposition}
    \label{prop:barygd_falminexist}
    Suppose at least one of the $\mu_i$'s is absolutely continuous.
    Then, for any $\al > 0$, the functional $F^\al$ admits at least one
    minimizer in $P_2((\RR^d)^n)$.
    Moreover, this minimizer is absolutely continuous with respect to the Lebesgue measure.
\end{proposition}

Given a minimizer $\gamma^\al$ of $F^\al$, the measure $\bary{\gamma^\al}$, dubbed the associated barycenter to $\gamma^\al$, 
should be close to the barycenter $b$ of $P$ thanks to \eqref{eqn:bary_pf} and \eqref{eqn:multmargform}.
Our next result quantifies this proximity with respect to $\al$ under extra assumptions on $\mu_1, \dots, \mu_n$.

We first assume that $\mu_1, \dots, \mu_n$ satisfy the Poincaré inequality with a strictly positive constant $\Ciso$, \ie that
for any $i \in \eint{1,n}$ and any Lipschitz $f \in L^2(\mu_i)$, we have
\begin{equation}
    \label{def:logsobolev}
    \nabs{f}_{L^2(\mu_i)}^2 \leq \Ciso \nabs{\nabla f}_{L^2(\mu_i)}^2
\end{equation}
where $\nabla f$ is defined Lebesgue-almost everywhere.
Such inequalities are very common in the sampling setting.
We also assume  that $P = \sum \lambda_i \delta_{\mu_i}$ satisfies a variance inequality with constant $k$,
\ie that there exists a strictly positive constant $k$ such that for any 
$\nu \in P_2(\RR^d)$, it holds
\begin{equation*}
    k W_2^2(\nu, b_P) \leq B(\nu) - B(b_P),
\end{equation*}
where $b$ is the barycenter of $\mu_1, \dots, \mu_n$, and $B : \nu \mapsto \sum \lambda_i W_2^2(\mu_i, \nu)$
is the functional appearing in the original formulation of the barycenter problem. 

Such inequalities were introduced by \cite{sturm_varineq} to describe curvature properties of geodesic spaces and have played
a central role in understanding the behavior of the empirical Wasserstein barycenter 
(see \cite{ahidar_coutrix_rates,legouicFastConvergenceEmpirical2019}). We refer to Appendix \ref{sect:proofs} for 
details on variance inequalities. 
We can now state the following result.

\begin{proposition}
    \label{prop:barygd_barstab}
    Let $\al \geq 1$ and $\varepsilon \leq 1$.
    Suppose $\mu_1, \dots, \mu_n$ satisfy a Poincar\'e inequality with constant $\Ciso$ and that 
    $\sum \lambda_i \delta_{\mu_i}$ satisfies a variance
    inequality with constant $\Cvar$. Then there exists a strictly positive constant $C$, only depending
    on the variance $\sigma^2 = \int c \Di\gamma^\star$ and $\Ciso$, such that for any $\geal \in P_2((\RR^d)^n)$
    satisfying $F^\al(\geal) - \inf F^\al \leq \epsilon$, it holds
    \begin{equation}
        \label{ieqn:barygd_stab_wbars_bound}
        \frac{\Cvar}{4} W_2^2(\bary{\geal}, \bary{\gamma^\star}) \leq \epsilon + \frac{C}{\sqrt{\al}}.
    \end{equation}
\end{proposition}

Proposition \ref{prop:barygd_barstab} shows that any approximate minimizer of $F^\al$ produces a barycenter
which distance to the desired barycenter $b$ is of order $O(1 / \sqrt{\al})$ and can therefore be controlled
to arbitrary precision with $\al$.

The fact that the minimization of $F^\al$ is carried out without constraints on the marginals of $\gamma$ allows 
to use classical optimization techniques.
The next section is devoted to the introduction of the gradient flow on the Wasserstein space of $F^\al$ that will be the 
backbone of our algorithm.
\subsection{Minimizing scheme}

We first briefly recall that the Wasserstein gradient of a functional $F : P_2(\RR^d) \to \RR$ at 
a point $\nu \in P_2(\RR^d)$ is a vector field on $\RR^d$ given by the Euclidean gradient of its first variation $f$, \ie
\begin{equation}
    \label{eqn:ngrad_W}
    \nabla_W F(\nu) := x\mapsto\nabla f(x) \in L^2(\RR^d,\RR^d;\nu)
\end{equation}
where $f$ is such that for any signed measure $\delta$ with $\int \!\Di\delta = 0$, we have
\begin{equation*}
    F(\nu + \epsilon \delta) = F(\nu) + \epsilon \int f \Di\delta + o(\epsilon).
\end{equation*}
For instance, the Wasserstein gradient of $\mu \mapsto \Penalization_{\nu}(\mu)$ is given by 
\[
    \nabla_W \Penalization_{\nu}(\mu) = 2 \nabla \frac{\mu}{\nu}.
\]
Intuitively, when a probability measure is seen as a large number of particles, the Wasserstein 
gradient of a function defines a vector field along which each particle should be moved in order to locally maximize the function.
More details about the differential calculus on the Wasserstein space are provided in Appendix \ref{sect:sosws}.

To sample from the barycenter of $P=\sum \lambda_i\, \delta_{\mu_i}$, we consider the flow described by
\begin{equation}
    \label{eq:algo_cont}
    \begin{cases}
        \partial_t X_t = - \nabla_W F^\al(\gamma(t))(X_t) \\
        X(0) = X_0 \sim \gamma(0),
    \end{cases}
\end{equation}
where $\gamma(t)$ denotes the distribution of $X_t$ at time $t > 0$.
Recall that for a measure $\gamma \in P_2((\RR^d)^n)$, we write $\gamma_i$ for its $i$-th marginal.
Using \eqref{eqn:ngrad_W}, the Wasserstein gradient of $F^\al$ is given by
\begin{equation*}
    \nabla_W F^\al(\gamma)(x)
        = \nabla c(x) + 2 \al \sum \lambda_i \nabla \frac{\gamma_i}{\mu_i}(x),
\end{equation*}
for any $\gamma\in P_2((\RR^d)^n)$ and $x\in\RR^d$.

The explicit Euler scheme for \eqref{eq:algo_cont} provides an approximation of $(X_{t_m})_m$ for $t_{m+1}-t_m=h_m$, with
\begin{equation}
    \label{eqn:bad_euler}
    X_{m+1} = X_m - h_m \left[\nabla c(X_m) + 2 \al \sum \lambda_i \nabla \frac{\gamma_i(t_m)}{\mu_i}(X_m)\right],
\end{equation}
where $\gamma(t_m)$ is the distribution of $X_m$ and $h_m$ is a given step size.

However, note that since $\gamma(t)$ is unknown, this discretization scheme is not implementable as it is.
In order to implement it, we use a kernel to approximate the Wasserstein gradient
of the penalization term in $F^\al$ as is done in \cite{liu_wang_svgd, chewi_legouic_lu_maunu_rigollet_lawgd}.
The next section provides more details.

\section{Implementation}
\label{sect:simulations}

As noted in Section~\ref{sect:algorithm}, Algorithm \eqref{eqn:bad_euler} cannot be implemented directly 
since there is no canonical way to recover the unknown distribution $\gamma(t_m)$ of the current 
state $X_{t_m}$ from the mere knowledge of $X_m$. In this section, we present a concrete implementation of
the scheme \eqref{eqn:bad_euler}.

\paragraph{The \ssf{SVGD} algorithm.}

We consider the popular \ssf{SVGD}
algorithm introduced in \cite{liu_wang_svgd} which uses an explicit kernel $\ksvgd$ independent
of the target measure.
The kernel $\ksvgd$ can be chosen to have a convenient
analytical form (\eg a Gaussian kernel) amenable to direct evaluation. Let
$\pi \propto e^{-V}$, with $V \in \prC^1(\RR^d)$, be the target measure.
\ssf{SVGD} aims to minimize the $\Penalization_\pi$ divergence by transporting randomly initialized points
along the flow described by
\begin{equation}
    \label{eqn:svgd_flow}
    \partial_t X_t = - \iosvgd_{\pi} \nabla \Penalization_{\pi}(\mu_t)(X_t),
\end{equation}
where $X_t \sim \mu_t$ and $\iosvgd_{\pi} : f \mapsto \int \ksvgd(\scdot, x) f(x) \Di\pi(x)$ is the integral operator
associated with $\ksvgd$. Note that integration by parts yields
\[
    \iosvgd_{\pi} \nabla \Penalization_{\pi}(\mu_t)(X_t)
        = \int \bb{\ksvgd(X_t, x) \, \nabla V(x)  - \nabla_2 \ksvgd(X_t, x)} \Di\mu_t(x).
\]
Approximating the integrals by averaging with samples $(X^i_t)_{i=1,\dots,N}$ distributed according to $\mu_t$, 
one gets the \ssf{SVGD} algorithm
\begin{equation}
    \label{eqn:svgd_iter}
    X_{t+1}^i = X_t^i +
        \frac{h_t}{N} \sum_{j=1}^N  \Big\{ -\ksvgd(X_t^i, X_t^j)\, \nabla V(X_t^j)
            + \nabla_2 \ksvgd(X_t^i, X_t^j) \Big\}
\end{equation}
where the initial points $X_0^1, \dots, X_0^N$ are chosen at random.

\paragraph{Proposed algorithm.}

To minimize $F^\al$, we implement \eqref{eqn:bad_euler}, replacing each gradient of 
the penalization sum, $\sum_{i=1}^n\lambda_i\Penalization_{\mu_i}$, with the \ssf{SVGD} iteration for each marginal.
Thus, we couple $n$ batches of $N$ points and generate $N$ samples approximately distributed according to the
barycenter $\bary{\gamma^\al}$ of a coupling $\gamma^\al$ minimizing $F^\al$. Our iteration takes the form
\begin{equation*}
    X_{t+1}^{i,j} 
        = X_t^{i,j} - h_t \nabla_j c(X_t) 
            + \frac{\al h_t \lambda_j}{N} \sum_{\ell=1}^N 
                \bb{ \ksvgd(X_t^{i,j}, X_t^{\ell,j}) \, \nabla \log \mu_j(X_t^{\ell,j}) + \nabla_2 \ksvgd(X_t^{i,j}, X_t^{\ell,j}) }
\end{equation*}
where $j \in \eint{1,n}$ is the index of marginal $\mu_j$ and $X_t^{i,j}$, $i \in \eint{1,N}$, stands for the 
$i$-th sample associated with the $j$-th marginal.

\paragraph{Numerical details.}

\begin{figure}[ht!]
    \centering
    \hspace*{-2em}
    \includegraphics[width=0.8\textwidth]{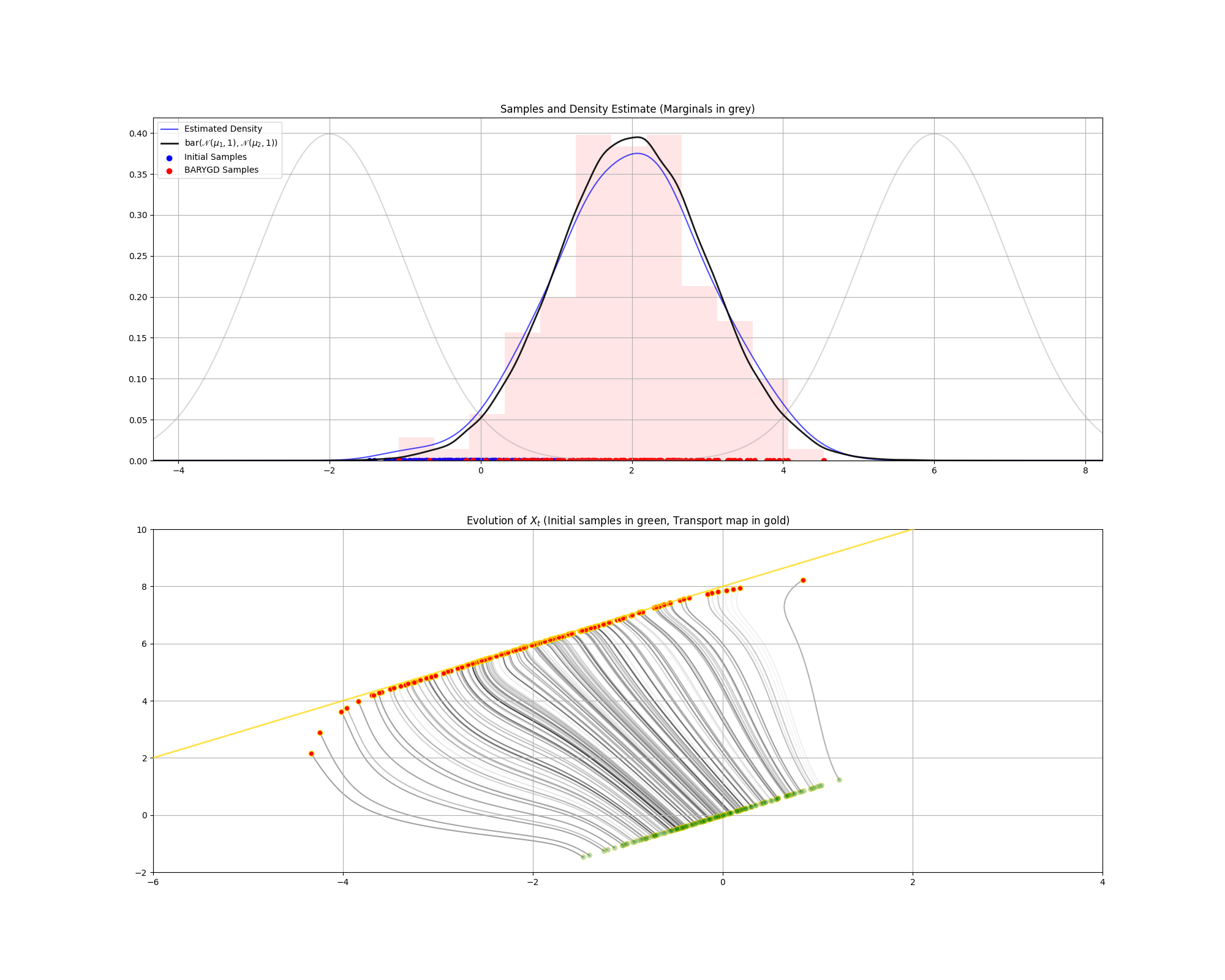}
    \includegraphics[width=0.8\textwidth]{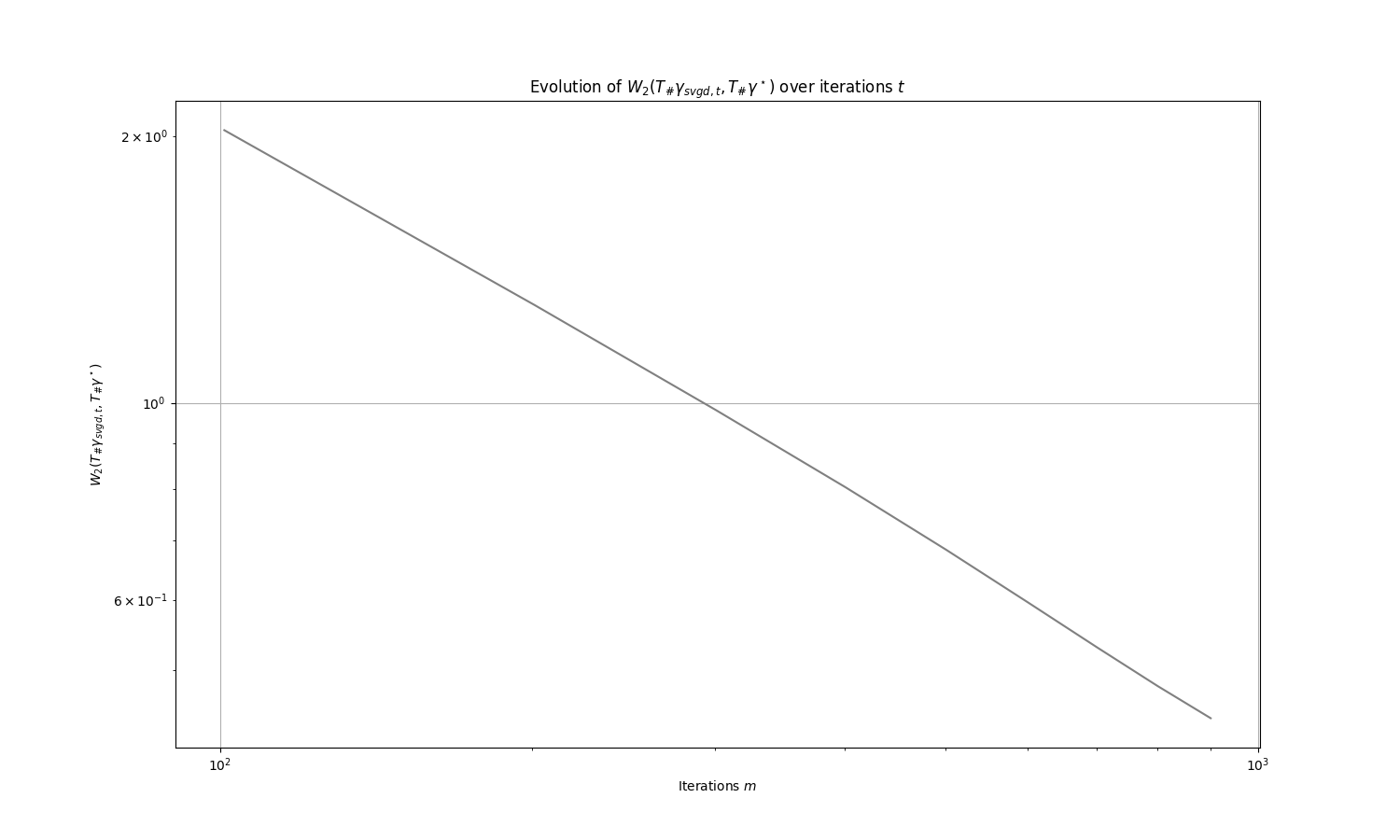}
    \caption{\small \textit{Top figure}: Top plot: 150 samples drawn from the barycenter of two unit variance normal
        distributions (shown in light gray) using our algorithm with Gaussian kernel with $\al = 10^3$ and $h_m = 10^{-4}$, 
        and the resulting kernel density estimator. 
        Bottom plot: Trajectories of the samples evolving over 1000 iterations, starting from green to red sample.
        The transport map is almost perfectly reconstructed.
        \textit{Bottom figure}: the Wasserstein distance to the true barycenter over iterations plotted in $\log-\log$ scale. 
        From the graph, we read that \ssf{BARYGD} performs as $O(1/\sqrt{m})$ in the number of iterations $m$.
    }
    \label{fig:norm}
\end{figure}

All the simulations are carried out with a Gaussian kernel 
\begin{equation*}
    \ksvgd(x,y) = \exp\bp{-\nabs{x-y}^2} \quad \forall x, y \in \RR.
\end{equation*}
The step size $h_t$ as well as $\al = \al_t$ vary over iterations, starting from $h_0=0.1$ 
and $\al_0=1000$. We first keep the parameters fixed and let gradient descent concentrate the particles on 
the graph of an increasing function before doubling the penalization strength while keeping $h_t \cdot \al_t$
constant. Thus, we enforce the marginal constraints once the coupling corresponds to an optimal transport.
In Figure \ref{fig:norm}, we the step-size evolves according to AdaGrad (see \cite{duchi_adagrad}).
We give further experimental results in Appendix \ref{sect:app_simuls}.

\section{Conclusion}
\label{sect:discussion}

We have presented a new algorithm to sample from the Wasserstein barycenter of a finite set of measures.
Our numerical experiments illustrate its convergence in several examples. 
We also proved a theoretical bound for samples drawn from almost minimizers of $F^\al$, the gradient 
flow of which inspired our algorithm.
It remains to formally prove that this algorithm, a kernelized gradient descent of $F^\al$, converges to a 
minimum of $F^\al$ and bound the rate of convergence.
On another note, it would be interesting to import other common tools of optimization to the Wasserstein setting
in order to derive a sampling algorithm by optimizing $F^\al$.

\titleformat{\section}[block]{\sffamily\textbf{\MakeUppercase{#1}}}{}{1em}{}

\section*{Acknowledgments}

We thank Philippe Rigollet for useful discussions.
Thibaut Le Gouic was supported by NSF award IIS-1838071.
\clearpage
\printbibliography

\newpage
\onecolumn
\appendix

\titleformat{\section}[block]{\centering\sffamily\textbf{\thesection. \, \MakeUppercase{#1}}}{}{1em}{}
 
\section{Additional Simulations}
\label{sect:app_simuls}

In this section, we illustrate our algorithm by other experiments. We also test the \ssf{LAWGD}
algorithm introduced by \cite{chewi_legouic_lu_maunu_rigollet_lawgd} to approximate the gradient of the penalization
in the Euler scheme \eqref{eqn:bad_euler}.

\paragraph{The \ssf{LAWGD} algorithm.}

\ssf{LAWGD} generates a batch of samples from the target measure $\pi \propto e^{-V}$ by transporting 
randomly initialized points $X_0^1, \dots, X_0^N \in \RR^d$ along the flow described by
\begin{equation}
    \label{eqn:lawgd_flow}
    \partial_t \mu_t = \div(\mu_t \nabla_2 \prL^{-1}(\mu_t))
\end{equation}
where $\prL = \Delta - \nabla V \scdot \nabla$ is the infinitesimal generator of a Markov diffusion process having 
stationary measure $\pi$. It is assumed to have discrete spectrum and that its 
eigenvectors $\psi_k$ (associated with strictly positive eigenvalues $\lambda_k$) form a basis of $L^2(\pi)$. An 
eigendecomposition then yields
\begin{equation}
    \label{eqn:k_lawgd}
    \prL^{-1} = \sum \frac{\psi_k \otimes \psi_k}{\lambda_k},
\end{equation}
which leads to the \ssf{LAWGD} algorithms upon discretization:
\begin{equation}
    \label{eqn:law_iter}
    X_{m+1}^i = X_m^i - \frac{h_m}{N} \sum_{j=1}^N \nabla_2 \prL^{-1}(X_m^i, X_m^j)
\end{equation}
where $h_m$ is the step size at iteration $m$. In contrast to \ssf{SVGD} where one uses $\nabla V$
directly, here the gradient intervenes only indirectly to compute the $\psi_k$
while the actual computations are carried out with $\nabla_2 \prL^{-1}$.
To compute eigendecomposition \eqref{eqn:k_lawgd}, we implement the Schrödinger scheme described in
\cite[Section 5]{chewi_legouic_lu_maunu_rigollet_lawgd}.

\paragraph{Barycenter of Gaussians.}

Let $b$ denote the Wasserstein barycenter of $\mu_i = \prN(m_i, S_i)$, $i = 1, \dots, n$.
We have $b = \prN(m^\star, S^\star)$ where 
$m^\star = \sum \lambda_i m_i$ and $S^\star$ is the fixed point
of $S \mapsto \sum \lambda_i (S^{1/2} S_i S^{1/2})^{1/2}$ on positive semidefinite
matrices. Moreover, the Wasserstein distance between two Gaussians is explicit:
\begin{equation*}
    W_2^2(\mu_i, \mu_j) = \nabs{m_i - m_j}^2 + \trace{S_i^2 + S_j^2 - 2(S_i S_j)^{1/2}}
\end{equation*}
for any $i, j \in \eint{1,n}$. See \cite{malago_riem_gaussians} for details on the geometry
of Gaussians and \cite[Section 6]{agueh_carlier_wbary} for the characterization of the barycenter of Gaussians.
These results allow us to compare our approximate barycenters to the true barycenters for
Gaussian marginals.

\paragraph{Experimental setup.}

\begin{figure}[ht!]
    \centering
    \includegraphics[scale=0.65]{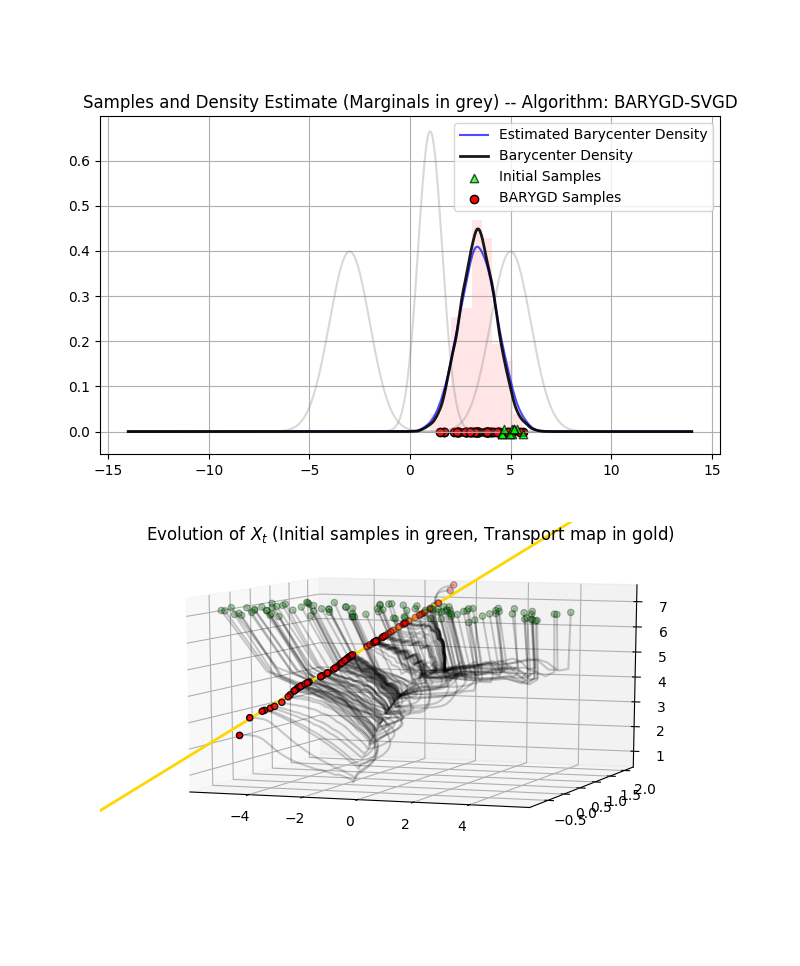}
    \caption{Top plot: 300 samples drawn from the barycenter of three normal distributions using
    \ssf{BARYGD} with a Gaussian kernel, and the density estimator. Initial samples are shown
    in light green. Bottom plot: Evolution of
    the samples over 600 iterations. We see the linear transport map is well reconstructed.
    The "front" is formed when the algorithm starts imposing the marginal constraints.}
    \label{fig:3_norms}
\end{figure}

\begin{figure}[ht!]
    \centering
    \includegraphics[scale=0.65]{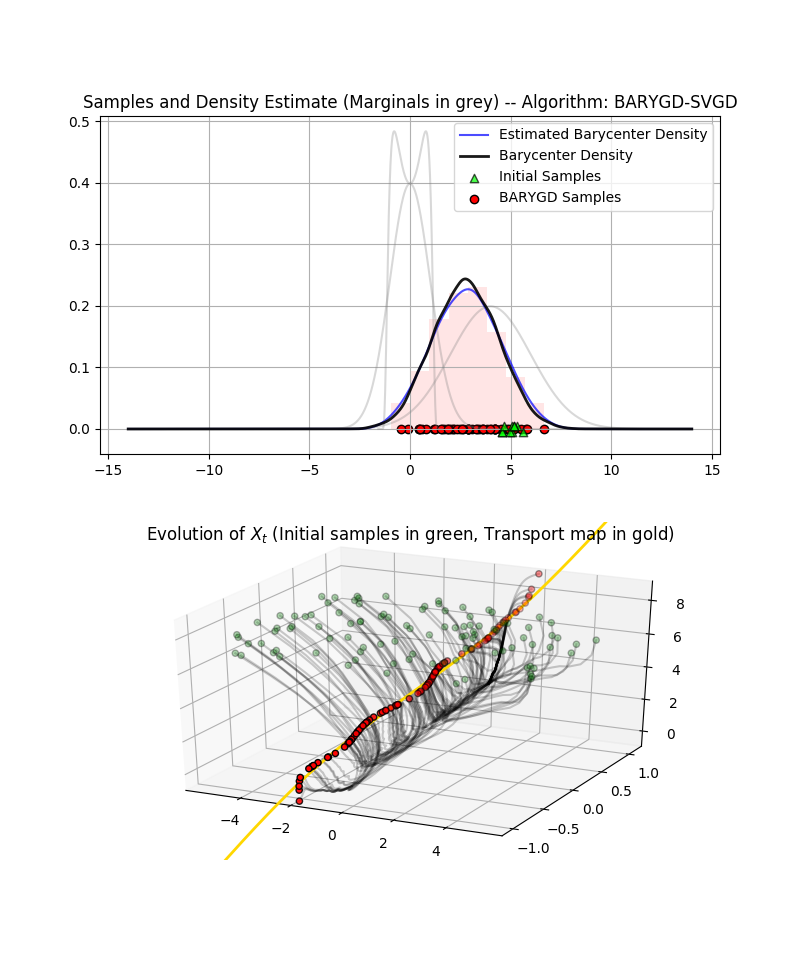}
    \caption{Top plot: 300 samples drawn from the barycenter of two normal distributions and 
    $\arctan \pf \prN(0,1)$ using
    \ssf{BARYGD} with a Gaussian kernel, and the density estimator. Initial samples are shown in
    light green. Bottom plot: Evolution of
    the samples over 600 iterations. We see the samples coil around the transport map, approximately
    reconstructing it. The "front" is formed when the algorithm starts imposing the marginal
    constraints.}
    \label{fig:3_norms_atan}
\end{figure}

In our experiments, we first consider testing \ssf{BARYGD} in one dimension with measures obtained by pushing 
forward the standard normal distribution by an increasing map. With three marginals, to choose the parameters 
$\al_t$ and $h_t$, we start with $h_0=0.1$ and $\al_0=1000$ and double $\al_t$ and divide $h_t$ by $2$---keeping
$\al_t\, h_t = \cte$---whenever all marginal samples are increasing functions of each other.
reference samples from the barycenter against which to compare our results.
Figures \ref{fig:norm_arctan} and \ref{fig:norm} illustrate our algorithm with two
marginals. Figures \ref{fig:3_norms} and \ref{fig:3_norms_atan} shows results with three marginals.
The golden line represents
the Gangbo-Święch map $T = (T^1, T^2, T^3)$ where $T^i$ are the coordinate transformations, \eg
$T^1 = \id, T^2 = \arctan, T^3 = (\textrm{an affine transformation})$ when the distributions are $\prN(0,1)$,
$\arctan \pf \prN(0,1)$ and $\prN(\mu, \sg^2)$. 

\begin{figure}[ht!]
    \centering
    \includegraphics[width=\textwidth, height=15cm]{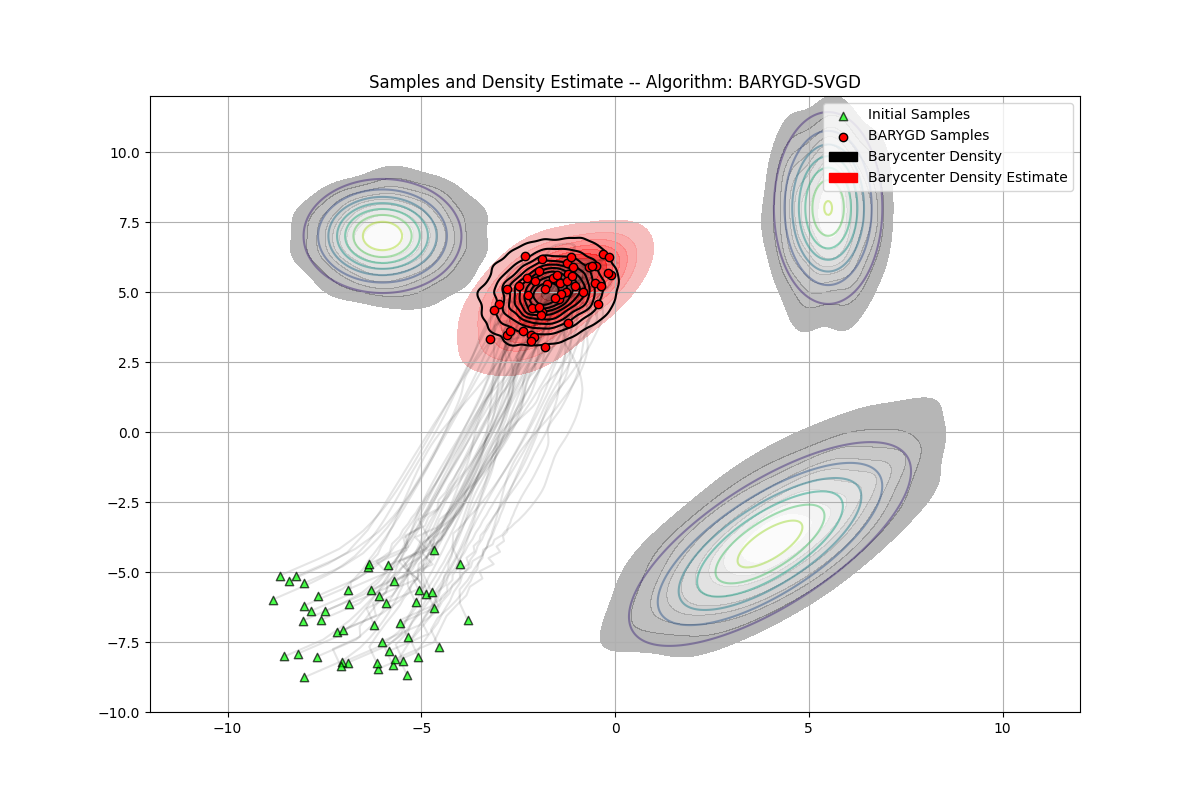}
    \caption{150 samples (red) drawn from the barycenter of three Gaussian distributions in the plane
    using \ssf{BARYGD} with an adaptive Gaussian kernel and adaptive step size (AdaGrad)
    over 1000 iterations. Initial points are shown in
    green. The colored regions represent kernel density estimates of the marginals computed based on the generated
    samples. Contour lines represent the reference measures.}
    \label{fig:2d-3-norms}
\end{figure}

Then we test the algorithm in two dimensions by sampling from the barycenter of three Gaussian
distributions. In this case, we choose $\al=1000$ and $h=10^{-4}$. This is shown in Figure 
\ref{fig:2d-3-norms}. The colored regions show the kernel estimates of the marginals. The contour 
lines represent the true marginals and barycenter.

\begin{figure}[ht!]
    \centering
    \includegraphics[width=\textwidth]{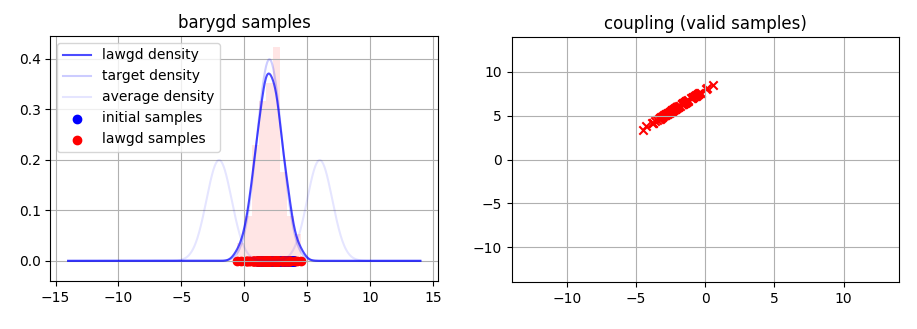}
    \caption{Left: 100 samples drawn from the barycenter of two Gaussian distributions with
    \ssf{BARYGD} with \ssf{LAWGD} kernel over 300 iterations. We see that the density is well approximated.
    Right: Coupling at the end of the algorithm. The points are aligned on a line, as
    expected when transporting Gaussians.}
    \label{fig:barygdlawgd}
\end{figure}

Choosing an appropriate sequence of step sizes that ensures gradient descent converges is a difficult
problem. We focused on testing \ssf{BARYGD} with \ssf{SVGD} as our implementation of this variant proved easier 
to parametrize. However, as illustrated on Figure \ref{fig:barygdlawgd}, we have observed that
when \ssf{BARYGD} with \ssf{LAWGD} kernel is well parametrized, it can yield qualitatively better results than when
\ssf{SVGD} is used.

\paragraph{Effect of the penalization strength.}

\begin{figure}[ht!]
    \centering 
    \begin{minipage}[b]{0.49\linewidth}
        \includegraphics[width=\linewidth]{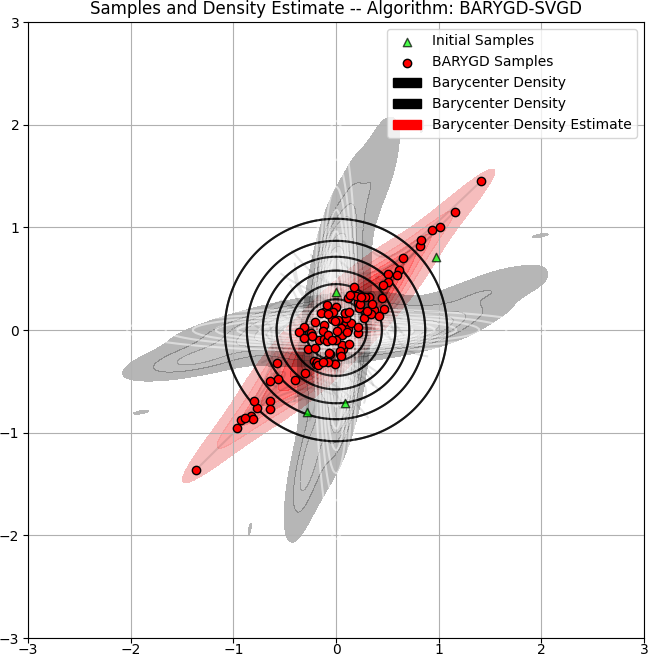}
    \end{minipage}
    \begin{minipage}[b]{0.49\linewidth}
        \includegraphics[width=\linewidth]{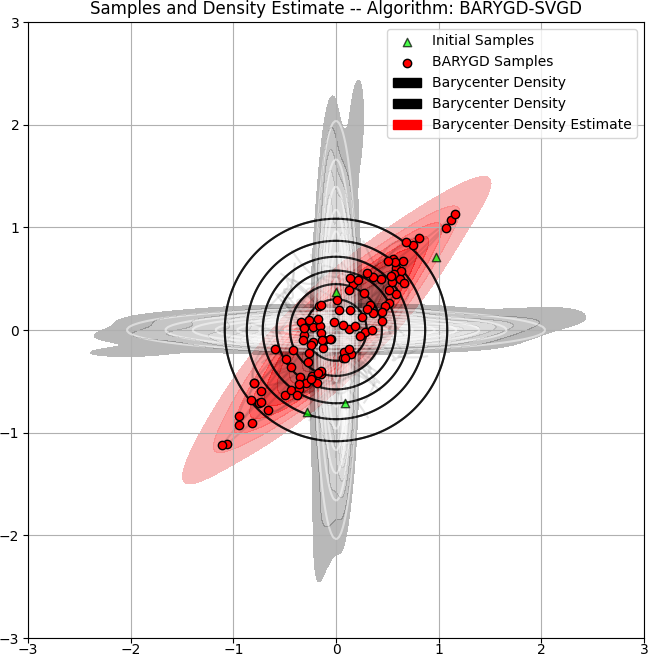}
    \end{minipage}
    \includegraphics[width=0.5\linewidth]{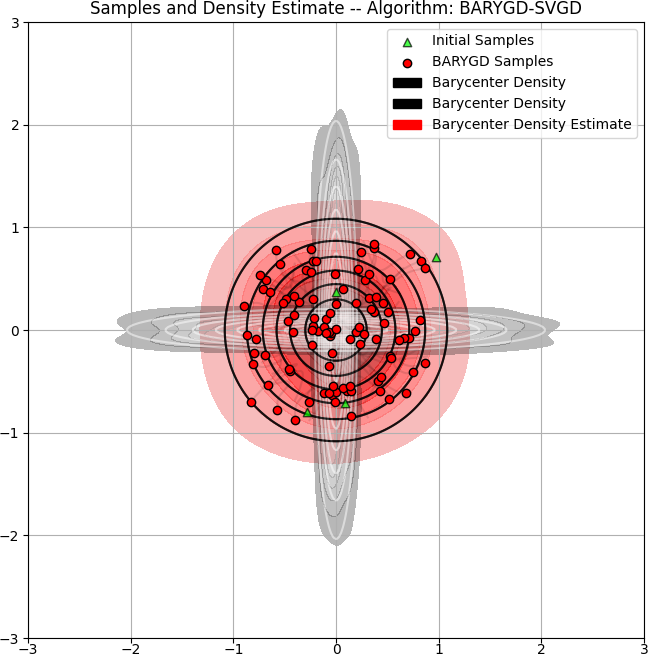}
    \caption{
        100 samples (red) drawn from the barycenter of two almost orthogonal normal distributions (gray patches)
        $\prN(0, \begin{bmatrix}1 & 0 \\ 0 & \epsilon\end{bmatrix})$ and 
        $\prN(0, \begin{bmatrix} \epsilon & 0 \\ 0 & 1\end{bmatrix})$ with $\epsilon = 10^{-2}$
        using \ssf{BARYGD}. We used a Gaussian kernel and fixed step size $h = 10^{-2} \al^{-1}$. The resulting
        density estimates are shown as red patches. Initial samples are shown in light green. The density of the 
        true barycenter is represented by the black contour lines and the densities of the marginals by the white
        contour lines. The algorithm ran for 2000 iterations with penalization strengths $\al = 1, 10^2, 10^3$ (top left to top right
        to bottom plots). 
        The barycentric weights are $\lambda_1 = \lambda_2 = 0.5$. We see that the marginals and the barycenter 
        are increasingly better approximated as the value of $\al$ increases.
    }
    \label{fig:ortho_norms_100regul}
\end{figure}

We illustrate the effect of the penalization coefficient, we generate samples distributed according to
the barycenter of two Gaussians in two dimensions (see Figure \ref{fig:ortho_norms_100regul}). In this
experiment, the step size and the penalization strength are kept fixed. We observe that the algorithm
the quality of the approximate barycenter increases with the penalization.

%
%

\clearpage

\section{First order differential structure in Wasserstein spaces}
\label{sect:sosws}

The fundamental elements upon which a Riemannian structure is constructed
are firstly the curves, from which the
tangent spaces are constructed, then the metric to give the tangent space an
Euclidean structure.  To do so on the Wasserstein space, it is helpful to think of a probability measure
as a vast collection of appropriately distributed particles in $\RR^d$. Then, properties of the measure can
be phrased in terms of properties of the particles and vice versa.

Let $\mu_0 \in P_2(\RR^d)$ and $X_0 \sim \mu_0$ be a
point on an integral curve $(X_t)_t$ of a vector field $(V_t)_t$, \ie $\partial_t X_t =
V_t(X_t)$, and let $\mu_t$ be the law of $X_t$. Differentiating $\mu_t$ by duality with
smooth, compactly supported functions $f \in \prC_c^\infty(\RR^d)$ yields
\begin{align}
    \db{\partial_t \mu_t, f}
        &= \Expect\, \partial_t f(X_t) \nonumber\\
        &= \Expect\, \db{\nabla f(X_t), V_t(X_t)}\nonumber\\
        &= \int_{\RR^d} \nabla f \cdot V_t \,\Di\mu_t \quad(= \db{\partial_t \mu_t, f}) \label{eq:tangent_intuition}\\
        &= -\int_{\RR^d} f\, \div(\mu_t V_t)).\nonumber
\end{align}
In particular $(\mu_t)_t$ satisfies the conservation of mass equation
\begin{equation}
    \label{eqn:conserv_mass}
    \partial_t \mu_t = -\div(\mu_t V_t)
\end{equation}
in a weak sense.
In comparison with the Riemannian analogue of \eqref{eq:tangent_intuition}, equality \eqref{eq:tangent_intuition} 
suggests to interpret  the tangent vector to the curve at time $t$ as the vector field $V_t$.
However, equation \eqref{eqn:conserv_mass} does not uniquely define such a vector as replacing $V_t$ with
$V_t + \hat{V}_t$ where $\div(\mu_t \hat{V}_t) = 0$, gives the same curve $\mu_t$ for a different "tangent vector".

Fortunately, optimal transport theory provides a natural choice for
$V_t$ when one wants to transport $\mu_t$ to a nearby
$\mu_{t+h}$ at minimal quadratic cost: gradients $\nabla (\psi_t - \half \nabs{\,\cdot\,}^2)$ of 
Kantorovich potentials $\psi_t$ --- \ie convex functions such
that $\mu_{t+h} = \nabla \psi_t \pf \mu_t$.
Indeed, the optimal trajectory
between $X_{t+h}$ and $X_t$ is a geodesic: $X_{t+\nu} = (1-\nu) X_t + \nu X_{t+h}$ for
$0 \leq \nu \leq 1$, thus
\[
    (\partial_\nu X_{t+\nu})_{| \nu = 0} = \nabla (-\half \nabs{X_t}^2 + \psi_t(X_t)).
\]
Motivated by this case, one then defines the tangent space at $\mu_t$ as
\[
    T_{\mu_t} P_2(\RR^d) 
        = \overline{\bb{\nabla \psi \,|\, \psi \in \prC^\infty(\RR^d)}}^{\,L^2(\mu_t)},
\]
equipped with the Hilbert structure inherited from $L^2(\RR^d,\RR^d; \mu_t)$.
We denote by $\db{\cdot,\cdot}_\mu$ such a scalar product given by $\db{f,g}_\mu=\int \db{f,g}\Di \mu$ for $f,g: \RR^d\to\RR^d$. 
This Hilbert structure is also consistent with the Benamou-Brenier formula for the Wasserstein metric
\begin{equation}
    \label{eqn:benamou_brenier}
    W_2^2(\mu_0, \mu_1)
        = \inf_{\mu \in \Gamma(\mu_0, \mu_1)} \int_0^1 \nabs{V_t}_{\mu_t}^2 \Di t
\end{equation}
where $\Gamma(\mu_0, \mu_1) = \bb{(\mu_t)_t \,|\, \mu(0) = \mu_0 \textrm{ and } \mu(1)=\mu_1, \partial_t\mu_t=-\div(\mu_t V_t)}$.
Equation (\ref{eqn:benamou_brenier}) is a perfect analogue of the length formula in Riemannian manifolds.
Moreover, it verifies that optimal transport curves are indeed the geodesics in the
Wasserstein space.

Now let $F : P_2(\RR^d) \to \RR$ be a functional on the Wasserstein space and let
$f$ denote its first variation $f$, \ie the function satisfying
\begin{equation*}
    (\partial_t F(\mu + t\rho))_{|t=0} = \int_{\RR^d} f \,\Di\rho
\end{equation*}
for signed measures $\rho$ with total mass $\int_{\RR^d} \Di \rho = 0$.
To define the gradient of $F$
at a point $\mu \in P_2(\RR^d)$, let $(\mu_t)_t$ be a curve such that $\mu_0 = \mu$. Then
formally $\mu_t = \mu_0 - t \,\div(\mu_0 v_0) + o(t)$ and
\begin{equation*}
    F(\mu_t)
        = F(\mu_0) - t \int_{\RR^d} f \,\div(\mu_0 V_0) + o(t)
\end{equation*}
so
\begin{equation*}
    (\partial_t F(\mu_t))_{|t = 0}
        = \db{\nabla_W F(\mu_0), V_0}_{\mu_0}
        = \db{\nabla f, V_0}_{\mu_0}.
\end{equation*}
This resembles the behavior of Riemannian gradients and suggests that we define the Wasserstein 
gradient of $F$ as the gradient of its first variation, \ie that we set
\begin{equation*}
    \nabla_W F(\mu) := \nabla f : \RR^d \to \RR^d\subset L^2(\RR^d,\RR^d,\mu)
\end{equation*}

With this machinery, functionals over the Wasserstein space can be
optimized by following Wasserstein gradient flows.
This was the main result of the seminal paper of Jordan, Kinderlehrer, and Otto \cite{jko}, 
in which they proved that the marginal distributions of Langevin diffusion forms a path in 
the Wasserstein space that is the gradient flow of the Kullback-Leibler divergence.

Let $\mu \in P_2^{ac}(\RR^d)$. For completeness, we compute the gradient of $\Penalization_\mu$ at 
$\nu \in P_2^{ac}(\RR^d)$. Let $\delta$ be a signed measure such that $\int \Di\delta = 0$. Then,
for small $t$, we have
\begin{equation*}
    \Penalization_\mu(\nu + t \delta)
        = \Penalization_\mu(\nu) + 2 t \int \bp{\frac{\nu}{\mu}-1} \delta + o(t)
        = \Penalization_\mu(\nu) + 2 t \int \frac{\nu}{\mu} \delta + o(t)
\end{equation*}
hence the first variation of $\Penalization_\mu$ at $\nu$ is $2\nu / \mu$, and
\begin{equation*}
    \nabla_W \Penalization_\mu(\nu) = 2 \nabla \frac{\nu}{\mu}.
\end{equation*}

\section{Proofs}
\label{sect:proofs}

In the next two sections, we recall some notions that will appear in the proofs.
We recall useful notation in the third section and then the different proofs are gathered.

\subsection{The \texorpdfstring{$\chi^2$}{chi-2} transportation inequality}

We recall a transportation inequality which connects the $\Penalization$ penalization with the Wasserstein
distance under the Poincaré inequality. We say that a measure $\mu \in P_2(\RR^d)$ satisfies a
$\Penalization$-transport inequality with positive constant $c$, denoted by $T_2^{\chi^2}(c)$, if we have
\begin{equation*}
    W_2^2(\mu, \nu) \leq 2c \, \Penalization_\mu(\nu)
\end{equation*}
for all $\nu \in P_2(\RR^d)$.
The following important result is due to \cite{ding_ledoux_ineq} (see also the note by \cite{ledoux_remarks_on_ineqs})
and makes the connection between the $\Penalization$ divergence and the Wasserstein distance precise.
\begin{proposition}[Ding]
    \label{prop:ding_ledoux}
    Let $\mu \in P_2(\RR^d)$ and $c \in \RR_+^\star$.
    If $\mu$ satisfies a Poincaré inequality with constant $c$ then it satisfies $T_2^{\chi^2}(2c)$.
    Conversely, if $\mu$ satisfies $T_2^{\chi^2}(c)$ then it satisfies a Poincar\'e inequality with constant $2c$.
\end{proposition}

\subsection{Variance inequalities}

Variance inequalities are crucial to the proof of Proposition \ref{prop:barygd_barstab} 
in Subsection \ref{ssect:stab}. Moreover, they naturally ensure uniqueness of the barycenter.
These inequalities were introduced by \cite{sturm_varineq} in his investigation of the curvature
of general metric spaces.
They have since played a central role in the study of
convergence rates of empirical barycenters in \cite{ahidar_coutrix_rates} 
and \cite{legouicFastConvergenceEmpirical2019}. Recently, \cite{chewi_bures} gave
a simple condition on the optimal transport from the barycenter to any other measure implying 
a variance inequality.

Recall that a probability measure
$P \in P_2(P_2(\RR^d))$ with barycenter $b_P \in P_2(\RR^d)$ satisfies a variance inequality with
constant $k \in [\,0,1\,]$ if
\begin{equation}\label{eq:varineq}
    k W_2^2(\nu, b_P)  \leq \int [ W_2^2(\mu,\nu) - W_2^2(\mu, b_P) ] \, P(\!\Di\mu)
\end{equation}
for any $\nu \in P_2(\RR^d)$. Note that such an inequality always holds with $k = 0$ and that
if $k > 0$ then the barycenter is unique. Variance inequalities express that
the variance functional $\nu \mapsto \int W_2^2(\mu, \nu) P(\!\Di\mu)$ behaves quadratically around
the barycenter $b_P$.

From \cite{sturm_varineq}, it is known that variance inequalities with constant $1$ are satisfied for 
probability measures defined over spaces of non positive Alexandrov curvature. The situation is more
contrasted for measures defined over spaces of non negative curvature, such as $P_2(\RR^d)$.
On such spaces,
variance inequalities may not hold for $k>0$. 
Let us recall a recent result in \cite[Theorem 6]{chewi_bures} in that direction.

Let $P \in P_2(P_2(\RR^d))$ and let $b_P \in P_2(\RR^d)$
 be a barycenter of $P$. 
 For any $\nu\in P_2(\RR^d)$, denote by $\phi_\nu$ the Kantorovitch potential from $b_P$ to $\nu$.
If there exists
$k: P_2(\RR^d) \to \RR^+$ such that $\phi_\nu$ is $k(\nu)$-strongly convex for $P$-almost any
$\nu \in P_2(\RR^d)$, and $\int \phi_\nu \Di P(\nu)=\id$, then $P$ satisfies a variance inequality with 
constant $\int k(\nu) P(\!\Di\nu)$.

\subsection{Notation}

We recall that $T$ denotes for the map $x \mapsto \sum \lambda_i x_i$, where the $\lambda_i$ are the barycentric 
weights, and that, for any $\al \geq 0$, we minimize the functional
\begin{equation*}
    F^\al : \gamma \longmapsto \int c \Di\gamma + \al \sum \lambda_i \Penalization_{\mu_i}(\gamma)
    \textrm{ where }
    c(x) = \sum \lambda_i \nabs{x_i - T(x)}^2 \quad \forall x \in (\RR^d)^n.
\end{equation*}
We write $\gamma^\star$ for the barycenter coupling for the $\mu_i$ with weights $\lambda_i$, \ie
it satisfies
\begin{equation}
    \label{eqn:gs_def}
    \bary{\gamma^\star} = \argmin_{\nu \in P_2(\RR^d)} \sum \lambda_i W_2^2(\mu_i, \nu).
\end{equation}
We assume throughout that the measures $\mu_1, \dots, \mu_n$ satisfy the Poincaré inequality
with constant $\Ciso$.

\subsection{Proof of Proposition \ref{prop:barygd_falminexist}}

Let $\al > 0$. We begin by showing that $F^\al$ is lower semi-continuous, \ie that for every convergent
sequence $(\gamma^k)$, with limit $\gamma$, we have
\begin{equation*}
    F^\al(\gamma) \leq \liminf_{k \to \infty} F^\al(\gamma^k).
\end{equation*}
Let $R = \sum \lambda_i \Penalization_{\mu_i}$ and $F : \gamma \longmapsto \int c \Di\gamma$ so that  $F^\al = F + \al R$. 
By \cite[Theorem 3.6]{polyanskiy_wu_infotheory}, $\al R$ is lower semi-continuous with respect to the 
weak topology on $P_2((\RR^d)^n)$ and it remains to show that this property also holds for $F$.
Let $m > 0$ and $c_m = c \land m$. Since $c$ is continuous, it follows that $c_m \in \prC_b^0((\RR^d)^n$, and
one sees that $c_m \geq 0$ is increasing in $m$ (\ie $c_m < c_{m'}$ when $m < m'$). Moreover, it is clear that
$c_m \converges{m \to \infty} c$ pointwise. For any $k \geq 0$
\begin{equation*}
    \int c_m \Di\gamma^k \leq \int c \Di\gamma^k
\end{equation*}
and taking the limit inferior in $k$ on both sides we obtain
\begin{equation*}
    \liminf_{k \to \infty} \int c_m \Di\gamma^k = \int c_m \Di\gamma
        \leq \liminf_{k \to \infty} \int c \Di\gamma^k
\end{equation*}
where the equality follows by weak convergence. We now use the monotone convergence theorem to obtain
\begin{equation*}
    \int c \Di\gamma \leq \liminf_{k \to \infty} \int c \Di\gamma^k
\end{equation*}
which shows that $F$ is lower semi-continuous with respect to the weak topology on $P_2((\RR^d)^n)$. 
Hence, $F^\al$ is lower semi-continuous.

Next, we show that there exists a minimizer for $F^\al$.
It is clear that $\inf F^\al \geq 0$. Let $(\gamma^k)$ be a minimizing sequence of probability 
measures on $P_2((\RR^d)^n)$, \ie it satisfies $F^\al(\gamma^k) \converges{k \to \infty} \inf F^\al$.
Then, from a certain rank $k_0$, for any $k\geq k_0$, we have $F^\al(\gamma^k) \leq \inf F^\al +1$.
Because for any coupling $\gamma$, we have 
\begin{equation*}
    F^\al(\gamma) 
        = \som{\gamma} + \sum \lambda_i \int (\nabs{T(x)}^2 - 2 \db{x_i, T(x)}) \gamma(\Di x)
            \,+ \al \sum \lambda_i R_{\mu_i}(\gamma),
        =: \som{\gamma} + r(\gamma),
\end{equation*}
and since the assumption implies that $r(\gamma^k) \leq \inf F^\al + 1$, it holds
\begin{equation}
    \label{ieqn:fal_bounded_moments}
    \som{\gamma^k} \leq 2 (\inf F^\al + 1) < \infty.
\end{equation}
This guarantees that the sequence $(\gamma^k)$ is tight. Indeed, if by contradiction $(\gamma^k)$ were 
not tight, then at least for some $i \in \eint{1, n}$ the sequence $(\gamma_i^k)$ is not tight either 
(if all the $\gamma_i$'s were tight, then, for any $\varepsilon>0$ there is a compact set $K_i^{\varepsilon}$ 
such that for all $k$ we have $\gamma_i^k(\RR^d \setminus K_i^{\varepsilon}) \leq \varepsilon/n$, and then 
$\gamma^k((\RR^d)^n \setminus \otimes \,K_i^{\varepsilon} ) \leq \varepsilon$, which would make 
$(\gamma^k)$ tight as well). Now if the sequence $(\gamma_i^k)$ is not tight, then for some $m>0$, for 
any compact set $K\subset \RR^d$ we have $\gamma_i^k(\RR^d\setminus K ) > m$, for a subsequence still denoted 
by $(\gamma_i^k)$, and then 
$\sum \int \nabs{x_i}^2 \Di\gamma_i^k(x_i) = m_2(\gamma^k) = \infty $, which contradicts
\eqref{ieqn:fal_bounded_moments}.
Thus $(\gamma^k)$ is tight. Then, by \cite[Lemma 6.14]{villani_oan}, at least a sub-sequence of $(\gamma^k)$
converges to some absolutely continuous $\gamma \in P_2((\RR^d)^n)$ (if $\gamma$ 
were to be singular, we would have $F^\al(\gamma) = \infty$ due to the penalization term, which
cannot be). The lower semi-continuity of $F^\al$ implies that $\gamma$ is a minimizer, which concludes the proof.

\subsection{Proof of Proposition \ref{prop:barygd_barstab}}
\label{ssect:stab}

\paragraph{Notation.}

We define the set of $\varepsilon$-approximate minimizers of $F^\al$
\begin{equation*}
    \AlmostMin = \bb{ \gamma \in P_2((\RR^d)^n) \,|\, 0 \leq F^\al(\gamma) - \inf F^\al \leq \varepsilon }.
\end{equation*}
where $\al$ and $\epsilon$ are strictly positive.

\subsubsection{Main result}

The main result of this subsection is Proposition \ref{prop:barygd_barstab}. It states that,
assuming the $\mu_i$ satisfy a variance inequality and a Poincaré inequality,
for any $\al \geq 1$ and a small $\varepsilon$, the distance between
barycenters of couplings in $\AlmostMin$ and the barycenter $\bary{\gamma^\star}$
is controlled with $\epsilon$ and $1 / \sqrt{\al}$. The implication of this result is that, in the 
regime $\epsilon \ll 1 \ll \al$ we are interested in, this distance is bounded by 
$\epsilon + 1/\sqrt{\al}$, up to a constant. We recall the precise statement.

\begin{namedthm*}{Proposition 2}
    Let $\al \geq 1$ and $\varepsilon \leq 1$.
    Suppose $\mu_1, \dots, \mu_n$ satisfy the Poincaré inequality with constant $\Ciso$ and that 
    $\sum \lambda_i \delta_{\mu_i}$ satisfies a variance
    inequality with constant $\Cvar$. Then there exists a constant $\Cstab$, only depending
    on the variance $\sigma^2 = \int c \Di\gamma^\star$ and $\Ciso$, such that for any $\geal \in \AlmostMin$,
    \begin{equation*}
        \frac{\Cvar}{4} W_2^2(\bary{\geal}, \bary{\gamma^\star}) \leq \epsilon + \frac{\Cstab}{\sqrt{\al}}.
    \end{equation*}
\end{namedthm*}

Let us now sketch the proof. For any $\geal \in \AlmostMin$, we have
\begin{equation}
    \label{ieqn:prop_stab_sketch_start}
    \frac{1}{2} W_2^2(\bary{\geal}, \bary{\gamma^\star})
        \leq W_2^2(\bary{\geal}, \bary{\bgeal}) + W_2^2(\bary{\bgeal}, \bary{\gamma^\star}),
\end{equation}
where $\bgeal$ denotes the coupling associated to the multimarginal problem for the marginals of $\geal$.
Our first step is to show that under the Poincaré inequality 
couplings in $\AlmostMin$ are close to $\Pi(\mu_1, \dots, \mu_n)$, \ie
their marginals are, up to a constant, $\sqrt{\epsilon/\al}$-close to the desired marginals; this is Lemma
\ref{lem:en_alep_marg_conv}. Assuming that $\sum \lambda_i \delta_{\mu_i}$ satisfies a variance inequality, 
we then use Lemmas \ref{lem:en_alep_marg_conv} and \ref{lem:en_barstab_intdiff} to derive a
bound on the distance between the barycenters of $\bgeal$ and $\gamma^\star$.

\begin{lemma}
    \label{lem:en_alep_marg_conv}
    Suppose $\mu_1, \dots, \mu_n$ satisfy the Poincaré inequality with constant $\Ciso$.
    Recall $\sigma^2=\int c \Di\gamma^\star$.
    Let $\al > 0$, $\varepsilon \leq \sigma^2$ and 
    let $\geal \in \AlmostMin$. Then
    \begin{equation}
        \label{eqn:crad}
        \sum \lambda_i W_2^2(\geal_i, \mu_i) \leq \frac{8 \Ciso \sg^2}{\al}
    \end{equation}
\end{lemma}

\begin{lemma}
    \label{lem:en_barstab_intdiff}
    Suppose the $\mu_i$ satisfy the $\Ciso$-Poincaré inequality. 
    Let $\al \geq \sigma^2$, $\varepsilon \leq \sigma^2$  and let $\geal \in \AlmostMin$. 
    Further, let $\bgeal$ be the barycenter coupling with weights $\lambda_1, \dots, \lambda_n$ 
    for $\geal_1, \dots, \geal_n$, and define
    \begin{equation*}
        \Cdelta = 4\sigma \bp{2\Ciso + \sqrt{2\Ciso}}.
    \end{equation*}
    Then
    \begin{equation}
        \label{ieqn:intdiff_ieqn1}
        \int c \Di\geal - \int c \Di\gamma^\star \leq \varepsilon,
    \end{equation}
    \begin{equation*}
        \int c \Di\bgeal - \int c \Di\gamma^\star 
            \leq 
                \frac{\Cdelta}{\sqrt{\al}},
    \end{equation*}
    and
    \begin{equation*}
        \int c \Di\geal - \int c \Di\bgeal \leq \varepsilon + 
        \frac{3\Cdelta}{\sqrt{\al}}
    \end{equation*}
\end{lemma}

Our second step is to show that a perturbed variance inequality holds for $\sum \lambda_i \delta_{\gamma_i}$ for any
$\gamma \in \AlmostMin$; this is Lemma \ref{lem:app_varineq}. We exploit this result to
derive a bound on the distance between the barycenters of $\geal$ and $\bgeal$. However, it
introduces a supplementary term (the second one in the right hand side of \eqref{ieqn:barygd_appvarineq_stm}), which
we need to control with $\varepsilon$ and $1/\sqrt{\al}$. We carry this out with Lemmas \ref{lem:en_alep_marg_conv} 
and \ref{lem:en_barstab_intdiff}. We conclude the proof by injecting the two distance bounds
back in \eqref{ieqn:prop_stab_sketch_start}.

\begin{lemma}
    \label{lem:app_varineq}
    Let $P, Q \in P_2(P_2(\RR^d))$. Assume that $P$ satisfies a $\Cvar$-variance inequality and that
    $P$ and $Q$ admit barycenters $b_P$ and $b_Q \in P_2(\RR^d)$. Let
    \begin{equation*}
        \sg^2(P) = \int W_2^2(\nu, b_P) \, P(\!\Di\nu)
        \eqand
        \sg^2(Q) = \int W_2^2(\nu, b_Q) \, Q(\!\Di\nu).
    \end{equation*}
    Then, for any $\mu \in P_2(\RR^d)$, 
    \begin{equation}
        \label{ieqn:barygd_appvarineq_stm}
        \frac{\Cvar}{2} W_2^2(\mu, b_Q)
            \leq 
                \int \bp{W_2^2(\nu, \mu) - W_2^2(\nu, b_Q)} Q(\!\Di\nu)  + C(\mu) \, W_2(Q, P)
    \end{equation}
    where
    \[
        C(\mu) = 2\sqrt{8[\sg^2(Q) + \sg^2(P) + W_2^2(b_P, b_Q)] + 2W_2^2(\mu, b_Q)} 
    \]
\end{lemma}

\subsubsection{Proofs}

\begin{proof}[\textbf{Proof of Lemma \ref{lem:en_alep_marg_conv}}]
    Let $\gamma^\al$ be a minimizer of $F^\al$. Let $R = \sum \lambda_i \Penalization_{\mu_i}$. The definition of 
    $\AlmostMin$ gives
    \begin{equation*}
        0 \leq \int c \Di\geal - \int c \Di\gamma^\al + \al R(\geal) - \al R(\gamma^\al) \leq \varepsilon,
    \end{equation*}
    whence
    \begin{equation*}
        \al R(\geal) \leq \varepsilon - \int c \Di\geal + \int c \Di\gamma^\al + \al R(\gamma^\al).
    \end{equation*}
    Since $\gamma^\al$ is minimizing, we have $F^\al(\gamma^\al) \leq F^\al(\gamma^\star)$ and,
    subtracting $\int c \Di\geal$ from both sides, we find
    \begin{equation*}
        -\int c \Di\geal + \int c \Di\gamma^\al + \al R(\gamma^\al) 
            \leq \int c \Di\gamma^\star - \int c \Di\gamma^\al 
            \leq \int c \Di\gamma^\star,
    \end{equation*}
    hence $\al R(\geal) \leq \varepsilon + \int c \Di\gamma^\star$. Now, 
    by Proposition \ref{prop:ding_ledoux}, the Poincaré inequality implies 
    $W_2^2(\mu_j, \scdot) \leq 4 \Ciso R_{\mu_j}$ for any $j \in \eint{1,n}$ and the assumption
    that $\epsilon \leq \int c \Di\gamma^\star = \sg^2$, thus
    \begin{equation*}
        \sum \lambda_i W_2^2(\geal_i, \mu_i) 
            \leq 4 \Ciso R(\geal)
            \leq \frac{4 \Ciso}{\al} \bp{\varepsilon + \int c \Di\gamma^\star}
            \leq \frac{8 \Ciso \sg^2}{\al},
    \end{equation*}
    which completes the proof.
\end{proof}

\begin{proof}[\textbf{Proof of Lemma \ref{lem:en_barstab_intdiff}}]\,

    \paragraph{(First inequality)}
    
    By definition,
    \begin{equation*}
        \int c \Di\geal - \int c \Di\gamma^\star
            = F^\al(\geal) - \al R(\geal) - F^\al(\gamma^\star)
            \leq F^\al(\geal) - F^\al(\gamma^\star)
    \end{equation*}
    and adding and subtracting $F^\al(\gamma^\al)$ yields
    \begin{equation*}
        F^\al(\geal) - F^\al(\gamma^\star)
            = \underbrace{\bb{F^\al(\geal) - F^\al(\gamma^\al)}}_{\leq \varepsilon \textrm{ since } \geal \in \AlmostMin} 
                + \underbrace{\bb{F^\al(\gamma^\al) - F^\al(\gamma^\star)}}_{
                    \leq 0 \textrm{ since } \gamma^\al \textrm{ minimizes } F^\al
                }
            \leq \varepsilon,
    \end{equation*}
    hence the inequality.
    
    \paragraph{(Second inequality)}
    
    By equality of the original and multimarginal formulations of the barycenter problem for $\bgeal$ and $\gamma^\star$,
    we have
    \begin{align*}
        \int c \Di\bgeal - \int c \Di\gamma^\star
            &= \sum \lambda_i\bp{W_2^2(\geal_i, \bary{\bgeal}) - W_2^2(\mu_i, \bary{\gamma^\star})} \nonumber \\
            &\leq \sum \lambda_i \bp{W_2^2(\geal_i, \bary{\gamma^\star}) - W_2^2(\mu_i, \bary{\gamma^\star})}
    \end{align*}
    where we used the fact that $\gamma^\star$ is not optimal for the $\geal_i$ to assert that 
    $\sum \lambda_i W_2^2(\geal_i, \bary{\bgeal}) \leq \sum \lambda_i W_2^2(\geal_i, \bary{\gamma^\star})$. 
    Combining the triangle inequality with the identity $a^2-b^2 = (a-b)(a+b)$ for all $a,b \in \RR$
    yields
    \begin{align*}
        \int c \Di\bgeal &\,- \int c \Di\gamma^\star \nonumber \\
            &\hspace{-1cm} \leq \sum \lambda_i(W_2(\geal_i, \bary{\gamma^\star}) - W_2(\mu_i, \bary{\gamma^\star})) 
                             (W_2(\mu_i, \bary{\gamma^\star}) + W_2(\geal_i, \bary{\gamma^\star})) \\
            &\hspace{-1cm} \leq \sum \lambda_i W_2(\geal_i, \mu_i) (W_2(\geal_i, \bary{\gamma^\star})
                + W_2(\mu_i, \bary{\gamma^\star})) \\
            &\hspace{-1cm} \leq \sum \lambda_i W_2(\geal_i, \mu_i) (W_2(\mu_i, \geal_i) 
                + 2 W_2(\mu_i, \bary{\gamma^\star})).
    \end{align*}
    Applying Cauchy-Schwarz inequality, we get
    \[
        \int c \Di\bgeal - \int c \Di\gamma^\star 
            \leq \sum \lambda_i W_2^2(\geal_i, \mu_i)
                + 2 \sqrt{\sum \lambda_i W_2^2(\geal_i, \mu_i)}
                    \sqrt{\sum \lambda_i W_2^2(\mu_i, \bary{\gamma^\star})}.
    \]
    We conclude with Lemma \ref{lem:en_alep_marg_conv}, which yields
    \begin{equation}
        \label{ieqn:intdiff_ieqn2}
        \int c \Di\geal - \int c \Di\gamma^\star
            \leq \frac{8 \Ciso \sg^2}{\al} + 2 \frac{\sg \sqrt{8 \Ciso}}{\sqrt{\al}} \sg
            \leq \frac{\Cdelta}{\sqrt{\al}}
    \end{equation}
    since we assume that $\sigma \geq \epsilon$.
    
    \paragraph{(Third inequality)}
    
    Adding and subtracting $\int c \Di\bgeal$ in \eqref{ieqn:intdiff_ieqn1}, we get
    \begin{equation*}
        \bb{\int c \Di\geal - \int c \Di\bgeal}
                + \bb{\int c \Di\bgeal - \int c \Di\gamma^\star} 
            \leq \varepsilon
    \end{equation*}
    and by moving the second term of the left hand side to the right hand side, we have
    \begin{equation}
        \label{ieqn:step3_lemintdiff}
        \int c \Di\geal - \int c \Di\bgeal
            \leq \varepsilon + \int c \Di\gamma^\star - \int c \Di\bgeal.
    \end{equation}
    Proceeding similarly as for \eqref{ieqn:intdiff_ieqn2}, we obtain using Lemma 
    \ref{lem:en_alep_marg_conv} that
    \begin{align*}
        \int c \Di\gamma^\star - \int c \Di\bgeal
            &\le \sum \lambda_i W_2^2(\geal_i, \mu_i)
                + 2 \sqrt{\sum \lambda_i W_2^2(\geal_i, \mu_i)}
                    \sqrt{\sum \lambda_i W_2^2(\geal_i, \bary{\bgeal})}
    \end{align*}
    However, by optimality of $\bgeal$ and using Inequality \eqref{ieqn:intdiff_ieqn2}
    and the assumption that $\al \ge \sigma^2$,
    \begin{equation*}
        \sum \lambda_i W_2^2(\geal_i, \bary{\bgeal})
            = \int c \Di\bgeal 
            \le \int c \Di\geal
            \leq \sigma^2 +\frac{8 \Ciso \sg^2}{\al}
                + 2 \frac{\sg \sqrt{8 \Ciso}}{\sqrt{\al}} \sigma 
            \le \left(\sigma + \sqrt{8 \Ciso} \right)^2.
    \end{equation*}
    Therefore, we proved
    \begin{align*}
        \int c \Di\gamma^\star - \int c \Di\bgeal 
            &\le \frac{8 \Ciso \sg^2}{\al} 
                + 2\frac{\sg \sqrt{8 \Ciso}}{\sqrt{\al}} 
                \left(\sigma + \sqrt{8 \Ciso} \right) \\
            &\leq \frac{1}{\sqrt{\al}} \bp{4 \sg(2 \Ciso + \sqrt{2 \Ciso}) + 16 \sg \Ciso}
            = \frac{4\sg}{\sqrt{\al}} \bp{6 \Ciso + \sqrt{2 \Ciso}}
            \leq \frac{3 \Cdelta}{\sqrt{\al}}
    \end{align*}
    which concludes the proof together with \eqref{ieqn:step3_lemintdiff}.
\end{proof}

\begin{proof}[\textbf{Proof of Lemma \ref{lem:app_varineq}}]

    Let $\mu \in P_2(\RR^d)$. From the variance inequality for $P$, we have
    \begin{align}
        \label{ieqn:lem15_decomp}
        \Cvar W_2^2(\mu, b_P)
            &\leq \int W_2^2(\mu, \nu) \, P(\!\Di\nu) - \int W_2^2(\nu, b_P) \, P(\!\Di\nu) \nonumber \\
            &= \int W_2^2(\mu, \nu) (P-Q)(\!\Di\nu)
                + \int W_2^2(\nu, b_Q) \, Q(\!\Di\nu) \nonumber \\
                &\hspace{1cm}- \int W_2^2(\nu, b_P) \, P(\!\Di\nu)
                + \int (W_2^2(\mu, \nu) - W_2^2(b_Q, \nu)) \, Q(\!\Di\nu) \nonumber \\
            &\leq \int W_2^2(\mu, \nu)(P-Q)(\!\Di\nu) + \int W_2^2(\nu, b_P)(Q-P)(\!\Di\nu) \nonumber \\
                &\hspace{4cm}+ \int (W_2^2(\mu, \nu) - W_2^2(\nu, b_Q)) \, Q(\!\Di\nu),
    \end{align}
    where we used the optimality of $b_Q$ as a barycenter for $Q$ in the last line. We
    start by bounding the first term in \eqref{ieqn:lem15_decomp}. Let $\Gamma$ be an optimal coupling 
    between $P$ and $Q$. Then
    \begin{align*}
        \int W_2^2(\mu, \nu) (P-Q)(\!\Di\nu)
            &= \int (W_2^2(\mu, \nu) - W_2^2(\mu, \zeta)) \,\Gamma(\!\Di\nu \Di\zeta) \\
            &= \int (W_2(\mu, \nu) - W_2(\mu, \zeta))(W_2(\mu, \nu) + W_2(\mu, \zeta)) \,\Gamma(\!\Di\nu \Di\zeta),
    \end{align*}
    and, from the triangle inequality, we have $W_2(\mu, \nu) \leq W_2(\mu, \zeta) + W_2(\zeta, \nu)$, hence
    \begin{equation*}
        \int W_2^2(\mu, \nu) (P-Q)(\!\Di\nu)
            \leq \int W_2(\nu, \zeta) (W_2(\mu, \nu) + W_2(\mu, \zeta)) \,\Gamma(\!\Di\nu, \Di\zeta).
    \end{equation*}
    Recall the notation
    \begin{equation*}
        \sg^2(\mu; P) = \int W_2^2(\nu, \mu) P(\!\Di\nu)
        \eqand
        \sg^2(\mu; Q) = \int W_2^2(\nu, \mu) \, Q(\!\Di\nu).
    \end{equation*}
    Applying Cauchy-Schwarz and Young inequalities,
    \begin{equation*}
        \int W_2^2(\mu,\nu)(P-Q)(\!\Di\nu) \leq W_2(Q, P) \sqrt{2(\sg^2(\mu;P) + \sg^2(\mu;Q))}.
    \end{equation*}
    Now, since variance inequality holds in the reverse direction for $k=1$ in nonnegatively curved space 
    (see \cite[Theorem 3.2]{ahidar_coutrix_rates}) 
    \begin{equation*}
        \sg^2(\mu; P) - \sg^2(b_P; P) \leq W_2^2(\mu, b_P)
        \eqand
        \sg^2(\mu; Q) - \sg^2(b_Q; P) \leq W_2^2(\mu, b_Q),
    \end{equation*}
    whence
    \begin{equation*}
        \int W_2^2(\nu, \mu)(P-Q)(\!\Di\nu)
            \leq W_2(Q, P) \sqrt{2(\sg^2(P) + \sg^2(Q) + W_2^2(\mu, b_P) + W_2^2(\mu, b_Q))}.
    \end{equation*}
    We follow similar steps to bound the second term of \eqref{ieqn:lem15_decomp} and obtain that
    \begin{equation*}
        \int W_2^2(\nu, b_P)(Q-P)(\!\Di\nu)
            \leq W_2(Q, P) \sqrt{2(\sg^2(Q) + \sg^2(P) + W_2^2(b_Q, b_P))}
    \end{equation*}
    Injecting these bounds into \eqref{ieqn:lem15_decomp} and recalling the definition
    of $A$, we have
    \begin{equation}
        \label{ieqn:lem15_prefinal}
        \Cvar W_2^2(\mu, b_P)
            \leq \int (W_2^2(\mu,\nu) - W_2^2(b_Q, \nu)) Q(\!\Di\nu) 
                + A(\mu; Q, P)\,W_2(Q, P).
    \end{equation}
    
    We conclude by applying \eqref{ieqn:lem15_prefinal} twice, with $\mu$ and $b_Q$, and the 
    triangle inequality to get
    \begin{align*}
        \frac{\Cvar}{2} W_2^2(\mu, b_Q)
            &\leq \Cvar W_2^2(b_Q, b_P) + k W_2^2(\mu, b_P) \\
            &\hspace{-1cm} \leq
                \Cvar W_2^2(b_Q, b_P) 
                    + A(\mu; Q, P) \, W_2(Q, P)
                    + \int (W_2^2(\nu, \mu) - W_2^2(\nu, b_Q)) \, Q(\Di\nu) \\
            &\hspace{-1cm} \leq
                \int (W_2^2(\mu, \nu) - W_2^2(b_Q, \nu)) \, Q(\!\Di\nu)
                    + W_2(Q, P) \bp{A(b_Q; Q, P) + A(\mu; Q, P)}.
    \end{align*}
\end{proof}

\begin{proof}[\textbf{Proof of Proposition \ref{prop:barygd_barstab}}]
    The proof follows the outline described above.
    Let $\geal \in \AlmostMin$ and let $\bgeal$ be the barycentric coupling for 
    $\geal_1, \dots, \geal_n$ with weights $\lambda_1, \dots, \lambda_n$. 
    Recall that $\Cdelta = 4\sigma\left(4\Ciso + \sqrt{2\Ciso}\right)$.
    From the triangle and Young's inequalities, we have
    \begin{equation}
        \label{ieqn:prop_proof_decomp}
        \frac{1}{2} W_2^2(\bary{\gamma^\star},\bary{\geal})
            \leq W_2^2( \bary{\gamma^\star},\bary{\bgeal}) + W_2^2(\bary{\bgeal},\bary{\geal} ), .
    \end{equation}
    
    \paragraph{(Step 1. Bound on $W_2( \bary{\gamma^\star},\bary{\bgeal})$)}
    
    We start by bounding the distance between the barycenters of $\bgeal$ and $\gamma^\star$. Using
    the variance inequality for $P$, we have
    \begin{align*}
        \Cvar W_2^2(\bary{\gamma^\star},\bary{\bgeal})
            &\leq \sum \lambda_i(W_2^2(\mu_i, \bary{\bgeal}) - W_2^2(\mu_i, \bary{\gamma^\star})) \\
            &\hspace{-2cm}= \sum \lambda_i (W_2(\mu_i, \bary{\bgeal}) - W_2(\mu_i, \bary{\gamma^\star}))
                         (W_2(\mu_i, \bary{\bgeal}) + W_2(\mu_i, \bary{\gamma^\star}))
    \end{align*}
    and, by the triangle applied to $W_2(\mu_i, \bary{\bgeal})$ in both factors, we get
    \begin{align}
        \label{ieqn:barstab_wbgeal_gs}
        \Cvar &\, W_2^2(\bary{\bgeal}, \bary{\gamma^\star}) \nonumber \\
            &\leq \sum \lambda_i (W_2(\mu_i, \gamma_i^\eal) + W_2(\gamma_i^\eal, \bary{\bgeal}) 
                - W_2(\mu_i, \bary{\gamma^\star})) \nonumber \\
                &\hspace{5cm} \scdot (W_2(\mu_i, \gamma_i^\eal) + W_2(\gamma_i^\eal, \bary{\bgeal}) 
                + W_2(\mu_i, \bary{\gamma^\star}))
                    \nonumber \\
            &= \sum \lambda_i W_2^2(\mu_i, \gamma_i^\eal)
                    + 2 \sum \lambda_i W_2(\mu_i, \gamma_i^\eal) W_2(\gamma_i^\eal, \bary{\bgeal}) \nonumber \\
                &\hspace{5cm} + \sum \lambda_i(W_2^2(\gamma_i^\eal, \bary{\bgeal}) - W_2^2(\mu_i, \bary{\gamma^\star})).
    \end{align}
    We now bound the second term of \eqref{ieqn:barstab_wbgeal_gs}.
    Using Cauchy-Schwarz the second term is bounded by
    \begin{align}
        \label{ieqn:barstab_crossterm_bound}
        \sum \lambda_i W_2(\mu_i, \geal_i) W_2(\geal_i, \bary{\bgeal})
            &\leq\sqrt{\sum \lambda_i W_2^2(\mu_i, \geal_i)}\sqrt{ \sum \lambda_i W_2^2(\geal_i, \bary{\bgeal})}.
    \end{align}
    Recall that $\sg^2(P) = \sum \lambda_i W_2^2(\mu_i, \bary{\gamma^\star})$ denotes the variance of $P$.
    By optimality of $\bary{\bgeal}$, using the triangle and Young inequalities yields
    \begin{align}
        \label{ieqn:barstab_jensen1}
        \sum \lambda_i W_2^2(\geal_i, \bary{\bgeal})
            = \sum \lambda_i W_2^2(\geal_i,\bary{\gamma^\star})
            &\leq  2\sum \lambda_i ( W_2^2(\geal_i, \mu_i) + W_2^2(\mu_i, \bary{\gamma^\star}))  \\
            &\overset{\textrm{(Lemma~\ref{lem:en_alep_marg_conv}})}{\leq}
                2 \left( \frac{8 \Ciso \sg^2}{\al} + \sg^2\right).
    \end{align}
    Therefore, using Lemma~\ref{lem:en_alep_marg_conv} again, and the assumption $\al\ge\sg^2$,
    \[
       \sum \lambda_i W_2(\mu_i, \geal_i) W_2(\geal_i, \bary{\bgeal}) 
            \leq \frac{\sg\sqrt{16 \Ciso}}{\sqrt{\al}} \sqrt{8 \Ciso + \sg^2}
    \]
    For the last term of \eqref{ieqn:barstab_wbgeal_gs}, the multimarginal formulation for $\bgeal$ 
    and $\gamma^\star$ shows that the last sum equals
    $\int c \Di\bgeal - \int c \Di\gamma^\star$, which we control by Lemma \ref{lem:en_barstab_intdiff}.
    The first term of \eqref{ieqn:barstab_wbgeal_gs} is bounded by 
    a direct application of Lemma \ref{lem:en_alep_marg_conv}.
    Therefore, we get from \eqref{ieqn:barstab_wbgeal_gs}:
    \begin{equation}
        \label{ieqn:stab_frag2}
        \Cvar W_2^2(\bary{\bgeal}, \bary{\gamma^\star})
            \le  \frac{\sg\sqrt{8 \Ciso}}{\sqrt{\al}}  
                + 2\frac{\sg\sqrt{16 \Ciso}}{\sqrt{\al}} \sqrt{8 \Ciso + \sg^2} 
                + \frac{\Cdelta}{\sqrt{\al}} \approx \frac{\Ciso\sigma^2}{\sqrt{\al}}
    \end{equation}

    \paragraph{(Step 2. Bound on $W_2(\bary{\geal}, \bary{\bgeal})$)}
    We first bound $W_2(\bary{\geal}, \bary{\bgeal})$ by applying Lemma~\ref{lem:app_varineq} with $\mu=\bary{\geal}$ and
    \begin{equation*}
        P := \sum \lambda_i \delta_{\mu_i} \eqand Q:=P^\eal := \sum \lambda_i \delta_{\geal_i}.
    \end{equation*}
    Since $P$ satisfies a $\Cvar$-variance inequality by assumption, this gives us
    \begin{equation}\label{eq:wvi}
    \frac{\Cvar}{2}W_2^2(\bary{\geal}, \bary{\bgeal}) \le \sum\lambda_i\bp{W_2^2(\geal_i, \bary{\geal}) - W_2^2(\geal_i, \bary{\bgeal_i})}   + C(\bary{\geal}) \, W_2(P^\eal, P)
    \end{equation}
    where   
    \[
        C(\bary{\geal}) = 2\sqrt{8[\sg^2(P^\eal) + \sg^2(P) + W_2^2(\bary{\gamma^\star}, \bary{\bgeal})] + 2W_2^2(\bary{\geal}, \bary{\bgeal})}.
    \]
    First, we bound $C(\bary{\geal})$.
    Using the multimarginal problem for $\bgeal$,
    the variance of $P^\eal$ is easily bounded with Lemma \ref{lem:en_barstab_intdiff}. Indeed, we have
    \begin{equation*}
        \sg^2(P^\eal)
            = \sum \lambda_i W_2^2(\gamma_i, \bary{\bgeal})
            = \int c \Di\bgeal
            \leq \sg^2 + \frac{\Cdelta}{\sqrt{\al}}
    \end{equation*}
    Then we bound $W_2^2(\bary{\geal}, \bary{\bgeal})$.
    Using again triangle and Young inequalities, we get the following bound by Lemma~\ref{lem:en_barstab_intdiff}
    \begin{align*}
        W_2^2(\bary{\geal}, \bary{\bgeal})
            &\le 2\sum\lambda_i W_2^2(\bary{\geal},\geal_i) + 2\sum\lambda_i W_2^2(\geal_i, \bary{\bgeal}) \\
            &\le 2\int c \Di\geal + 2\int c\Di\bgeal\\
            &\le 2\varepsilon + 6\frac{\Cdelta}{\sqrt{\al}} + 4\sg^2(P^\eal)\\
            &\le 6\frac{\Cdelta}{\sqrt{\al}} + 6\sg^2 + 4\frac{\Cdelta}{\sqrt{\al}}.
    \end{align*}
    Since, $\sg^2=\sg^2(P) = \int c \Di\gamma^\star$, it follows from gathering our bounds and 
    using \eqref{ieqn:stab_frag2} that
    \begin{align}
        \label{ieqn:barstab_Abgeal}
        & C(\bary{\bgeal}) \nonumber \\
            &\leq 2 \sqrt{8\left[ 2\sg^2 + \frac{\sg\sqrt{8 \Ciso}}{\sqrt{\al}}  
                + 2\frac{\sg\sqrt{16 \Ciso}}{\sqrt{\al}} \sqrt{8 \Ciso + \sg^2} 
                + \frac{\Cdelta}{\sqrt{\al}} \right] + 12\frac{\Cdelta}{\sqrt{\al}} + 12\sg^2 +
                    8\frac{\Cdelta}{\sqrt{\al}} 
                    } \nonumber \\
            &= 2\sqrt{28\sg^2 + 21 \frac{\Cdelta}{\sqrt{\al}} 
                + 2\frac{\sg \sqrt{16 \Ciso}}{\sqrt{\al}} \sqrt{8 \Ciso + \sg^2}}
            =:A^b \approx \sqrt{\frac{\Ciso}{\sqrt{\al}}+\sg^2}.
    \end{align}
    Remark that $W_2^2(P,P^\eal)\le \sum\lambda_i W_2^2(\mu_i,\geal_i)$.
    Collecting our bounds, we can now apply \eqref{eq:wvi} to bound $W_2^2(\bary{\geal}, \bary{\bgeal})$ 
    by using Lemma~\ref{lem:en_barstab_intdiff}
    \begin{align}
        \label{ieqn:stab_frag1}
        \frac{\Cvar}{2} & W_2^2(\bary{\geal}, \bary{\bgeal}) \nonumber \\
            &\leq \sum \lambda_i \bp{ W_2^2(\geal_i, \bary{\geal}) 
                - W_2^2(\geal_i, \bary{\bgeal}) } + A^b W_2(P, P^\eal) \nonumber\\
            &\leq \int c \Di\geal - \int c \Di\bgeal
                + A^b \sqrt{\sum \lambda_i W_2^2(\geal_i, \mu_i)} \nonumber \\
            &\leq \varepsilon + \frac{1}{\sqrt{\al}} \bp{
                    3\Cdelta +  \sg A^b \sqrt{8 \Ciso}  }.
    \end{align}
    
    \paragraph{(Step 3. Conclusion)}
    We conclude by applying the bounds \eqref{ieqn:stab_frag1} and \eqref{ieqn:stab_frag2} 
    to \eqref{ieqn:prop_proof_decomp} to obtain
    \[
        \frac{k}{2} W_2^2(\bary{\gamma^\star},\bary{\geal})
            \le 2\varepsilon + \frac{\sg\sqrt{8 \Ciso}}{\sqrt{\al}}  
                + 2\frac{\sg\sqrt{16 \Ciso}}{\sqrt{\al}} \sqrt{8 \Ciso + \sg^2} 
                    + \frac{\Cdelta}{\sqrt{\al}} + \frac{2}{\sqrt{\al}} \bp{
                        3\Cdelta +  \sg A^b \sqrt{8 \Ciso}  }.
    \]
    The proof is complete setting
    \begin{equation}
        \label{ieqn:def_cstab}
        2 \Cstab
            = \sg \sqrt{8 \Ciso}
                + 2\sg \sqrt{16 \Ciso(8 \Ciso + \sg^2)} 
                + \Cdelta+ 2\bp{ 3\Cdelta +  \sg A^b \sqrt{8 \Ciso}  }.
    \end{equation}
\end{proof}
\end{document}